# EA-LiteUNet: An Edge-Adaptive and Resource-Efficient U-Net for Boundary-Sensitive Dermoscopic Image Segmentation

Wang Jiangtao[1,2], Nur Intan Raihana Ruhaiyem[*,2], Fu Panpan[2], Yang Yu[2], Huang Yan[2]

*1 School of Network Communication, Zhejiang Yuexiu University, Shaoxing,312000, China*

*2 School of Computer Sciences, Universiti Sains Malaysia, Penang,11800, Malaysia*

Corresponding Author Email: intanraihana@usm.my

**Abstract**:

Accurate boundary delineation remains a persistent challenge in dermoscopic image segmentation because of blurred lesion margins, heterogeneous textures, and complex background artifacts. From a signal-processing perspective, lesion boundaries represent high-frequency components that are highly susceptible to aliasing, noise amplification, and information loss. Consequently, repeated downsampling and feature transformations in conventional convolutional architectures often lead to severely degraded boundary representations. To address these limitations, we propose EA-LiteUNet, an edge-adaptive and computationally efficient U-Net variant specifically designed for boundary-sensitive medical image segmentation. The architecture integrates three core mechanisms: (1) boundary-aware representation learning to suppress aliasing and preserve high-frequency structural details; (2) attention-guided feature modulation to selectively enhance boundary-relevant responses across multi-scale features; and (3) a resource-adaptive inference strategy to dynamically balance segmentation accuracy and computational efficiency. Extensive evaluations across three public dermoscopic datasets demonstrate that EA-LiteUNet consistently achieves superior boundary precision. Specifically, on the ISIC 2018 dataset, the method significantly reduces the 95% Hausdorff Distance (HD95) to 12.89 pixels while maintaining a robust Dice score of 92.08%. Notably, this strong performance is achieved with an ultralightweight configuration of merely 0.29M parameters and 1.17 GFLOPs. Ablation studies further validate the complementary effects of these components, confirming their contribution to enhanced boundary fidelity and stable optimization.

## 1. Introduction

Accurate biomedical image segmentation plays a crucial role in medical image analysis and computer-assisted diagnosis. In dermoscopic imaging, segmentation accuracy largely depends on precise boundary delineation, as lesion margins often exhibit blurred transitions, irregular shapes, and heterogeneous textures. Recent studies indicate that boundary regions remain a dominant source of segmentation errors, even in advanced deep learning frameworks, highlighting the ongoing difficulty of robust boundary modeling in biomedical segmentation tasks[1,2].

From a signal-processing standpoint, object boundaries are primarily represented by high-frequency components that are sensitive to noise, spatial resolution reduction, and nonlinear feature transformations. During multi-scale feature learning, repeated downsampling without appropriate low-pass filtering may introduce aliasing artifacts and structural distortions, leading to degraded boundary representation [3,4]. Such degradation is not always reflected by region-based metrics, which may yield high Dice scores despite discontinuous contours or inaccurate boundary localization [5].

The U-Net architecture and its variants have been widely adopted backbones for biomedical image segmentation because of their encoder–decoder design and effective multi-scale feature aggregation. Empirical studies have shown that well-optimized U-Net-based models remain competitive across diverse biomedical datasets [1,7]. However, several limitations persist. Conventional downsampling operations may distort high-frequency boundary information, whereas standard skip connections typically transfer encoder features to the decoder without selective filtering, allowing boundary-irrelevant or noisy responses to interfere with contour reconstruction[8]. To improve boundary sensitivity, recent works have incorporated attention mechanisms, selective feature fusion, and boundary-guided supervision strategies[9,10,11]. Although these approaches enhance segmentation performance, they often increase computational complexity or rely on heuristic fusion schemes without explicitly regulating boundary-related signal responses throughout the network. In addition, most existing architectures employ fixed inference pathways, thereby limiting their adaptability under varying signal complexity and computational constraints—an important consideration for practical clinical deployment[12,13].

In this work, we propose EA-LiteUNet, a boundary-adaptive and computationally efficient U-Net variant that integrates boundary-aware representation learning with resource-adaptive inference within a unified framework. Specifically, a boundary-adaptive encoding mechanism is introduced to preserve high-frequency structural information during downsampling and mitigate aliasing-induced distortion. Boundary-aware representations are further embedded into skip connections to selectively enhance informative contour features while suppressing background interference. Moreover, a resource-adaptive inference module is incorporated into the decoder to dynamically adjust computational depth according to the signal complexity and resource availability. Through these designs, EA-LiteUNet aims to improve boundary fidelity and segmentation robustness while maintaining lightweight efficiency.

The main contributions of this work are summarized as follows:

(i) We design a lightweight boundary-adaptive encoding strategy that explicitly regulates high-frequency boundary-related signals during downsampling and feature extraction, thereby reducing aliasing-induced distortions and preserving fine-grained structural continuity.

(ii) We introduce a boundary-aware feature modulation mechanism in skip connections, enabling the selective enhancement of informative contour responses while suppressing boundary-irrelevant activations during multi-scale fusion.

(iii) A resource-adaptive inference mechanism is incorporated into the decoder to regulate the refinement depth based on boundary-related uncertainty, enabling an effective balance between the segmentation accuracy and computational complexity under varying signal conditions.

To further illustrate the practical impact of the proposed adaptive design, we analyze the boundary–efficiency trade-off of EA-LiteUNet on the ISIC2018 dataset, as shown in Fig. 1. The visualization demonstrates that the proposed framework achieves a lower boundary discrepancy while maintaining a compact parameter size and a moderate computational cost. This observation motivates the subsequent quantitative and ablation analyses presented in the experimental section.

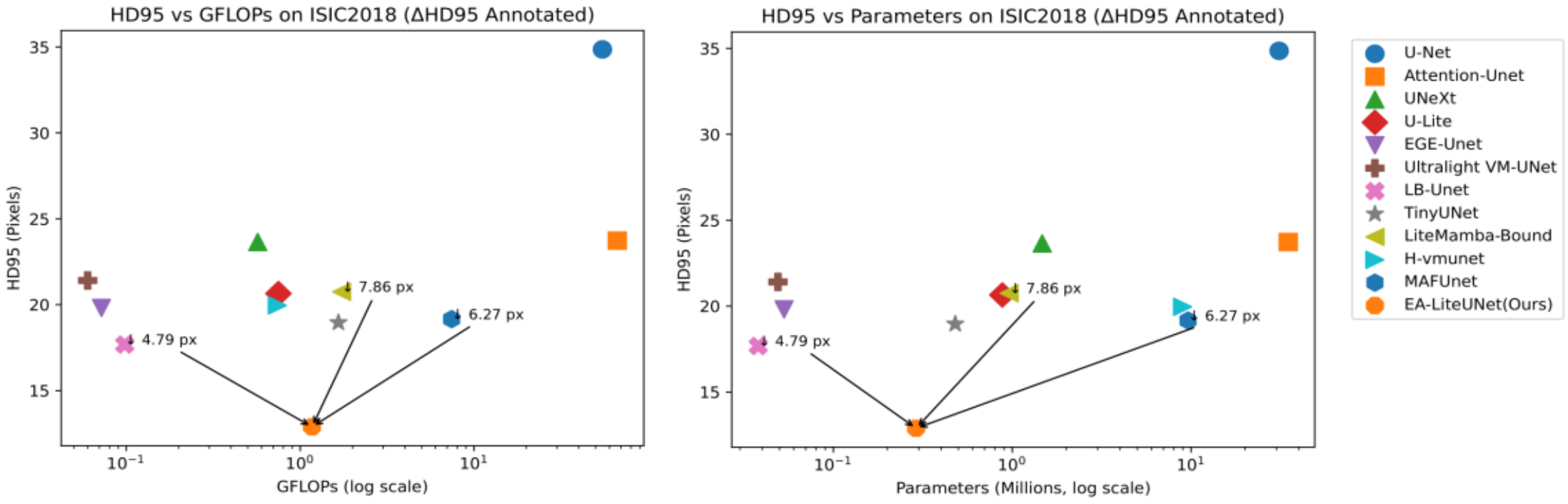

**Fig. 1.** Boundary-efficiency trade-off on ISIC2018 with ΔHD95 annotations. Illustrative Trade-off between boundary segmentation performance, Hausdorff Distance (HD95), and computational cost for representative dermoscopic image segmentation models on the ISIC2018 dataset. The left panel plots HD95 against Giga Floating Point Operations (GFLOPs), and the right panel plots HD95 against model parameters (M). The ideal model occupies the bottom-left region (high accuracy, low cost).

## 2. Related Work

Recent advances in deep learning have substantially improved the performance of dermoscopic image segmentation. Nevertheless, achieving accurate boundary delineation under constrained computational resources remains a persistent challenge. In this section, we review prior work from three primary perspectives: boundary-oriented segmentation strategies, attention-guided feature modulation in U-Net variants, and efficient or adaptive inference mechanisms.

### 2.1 Boundary-Oriented Segmentation Strategies

Accurate boundary modeling is critical in medical image segmentation, particularly for irregular lesions and low-contrast contours. To enhance boundary precision, numerous studies have incorporated auxiliary boundary supervision[14], boundary-sensitive loss functions[15], and contour-guided feature extraction mechanisms[16]. For instance, explicit boundary loss terms are frequently introduced to penalize contour discrepancies, a strategy that is especially beneficial in class-imbalanced scenarios[10,14]. Other approaches integrate edge-detection branches or contour-prediction modules to provide supplementary supervisory signals during training.

Although these methods successfully enhance boundary awareness at the supervision level, they operate predominantly at the output stage. Consequently, the degradation of boundary-related information during early feature extraction and downsampling is often left unaddressed. Because boundary structures inherently correspond to high-frequency image components, repeated downsampling without adequate signal preservation can introduce severe aliasing artifacts and structural distortion early in the encoding process[3, 17]. Such distortions can propagate through the network, limiting the effectiveness of supervision-driven boundary enhancement.

### 2.2 Attention-Guided Feature Modulation in Lightweight U-Net Variants

Attention mechanisms have been widely adopted in U-Net-based architectures to enhance feature selectivity and overall segmentation accuracy. Channel, spatial, and hybrid attention modules are typically employed to emphasize informative features while suppressing irrelevant background noise[18,19,20.] Specifically, attention-guided skip connections have been utilized to regulate encoder-decoder information flow and mitigate the semantic mismatch between hierarchical features[21, 22].

Despite their demonstrated effectiveness, most existing attention modules operate within a static feature domain, and fail to explicitly incorporate boundary uncertainty or evolving prediction cues during the decoding phase. Furthermore, conventional attention mechanisms frequently introduce substantial parameter counts and computational overhead, diminishing their suitability for resource-constrained clinical environments. These limitations point to a need for lightweight, prediction-aware modulation strategies that explicitly target boundary-sensitive feature refinement[23].

### 2.3 Efficient and Adaptive Inference Mechanisms

Driven by the growing demand for real-time, resource-aware clinical deployment, computational efficiency has emerged as a central concern in medical image analysis. Recent optimization efforts have focused on lightweight U-Net variants, efficient convolutional operations, and compact architectural designs[24]. Furthermore, adaptive inference mechanisms—such as early-exit strategies[25] and dynamic depth control[8, 25, 26]—have been explored to dynamically allocate computational resources based on input complexity.

Although these approaches effectively reduce computational costs, most focus on global efficiency optimization without explicitly accounting for localized boundary complexity. In practical clinical scenarios, boundary complexity varies significantly across images and within distinct spatial regions of the same image. Consequently, relying on fixed inference pathways often leads to either redundant computations on simple regions or insufficient contour refinement on complex margins. Currently, only a limited number of studies have investigated adaptive inference strategies explicitly tailored to boundary-sensitive segmentation[12, 27], revealing a distinct gap between efficiency-oriented design and boundary-aware modeling.

### 2.4 Summary and Motivation

In summary, existing boundary-aware methods primarily emphasize supervision and loss design, whereas attention-based approaches generally lack explicit awareness of boundaries and predictions. Conversely, efficiency-driven strategies rarely incorporate localized boundary complexity into their inference decisions. These observations point to a gap that has received comparatively little attention: few existing methods jointly address boundary-aware representation learning, selective feature modulation, and resource-adaptive inference within a single framework. Motivated by these compounded limitations, we propose an edge-adaptive and computationally efficient architecture specifically engineered to enhance boundary-sensitive segmentation within practical resource constraints.

## 3. Method

### 3.1 Overall Architecture

The proposed EA-LiteUNet adopts a lightweight U-shaped encoder–decoder architecture as its backbone, a paradigm widely validated for medical image segmentation because of its effective multi-scale representation and spatial detail recovery capability[1,28]. The details of the overall architecture are shown in Fig. 2. To enhance boundary-sensitive representation learning under strict computational constraints, the architecture incorporates three key components: (i) a stage-specific encoder equipped with an efficient boundary-aware representation module (EBARM), (ii) attention-guided skip connections for selective feature transfer, and (iii) a boundary-aware resource-adaptive inference mechanism in the decoder. The encoder consists of six stages with channel configurations {8, 16, 24, 32, 48, 64}, ensuring progressive representational growth while maintaining a compact model footprint.

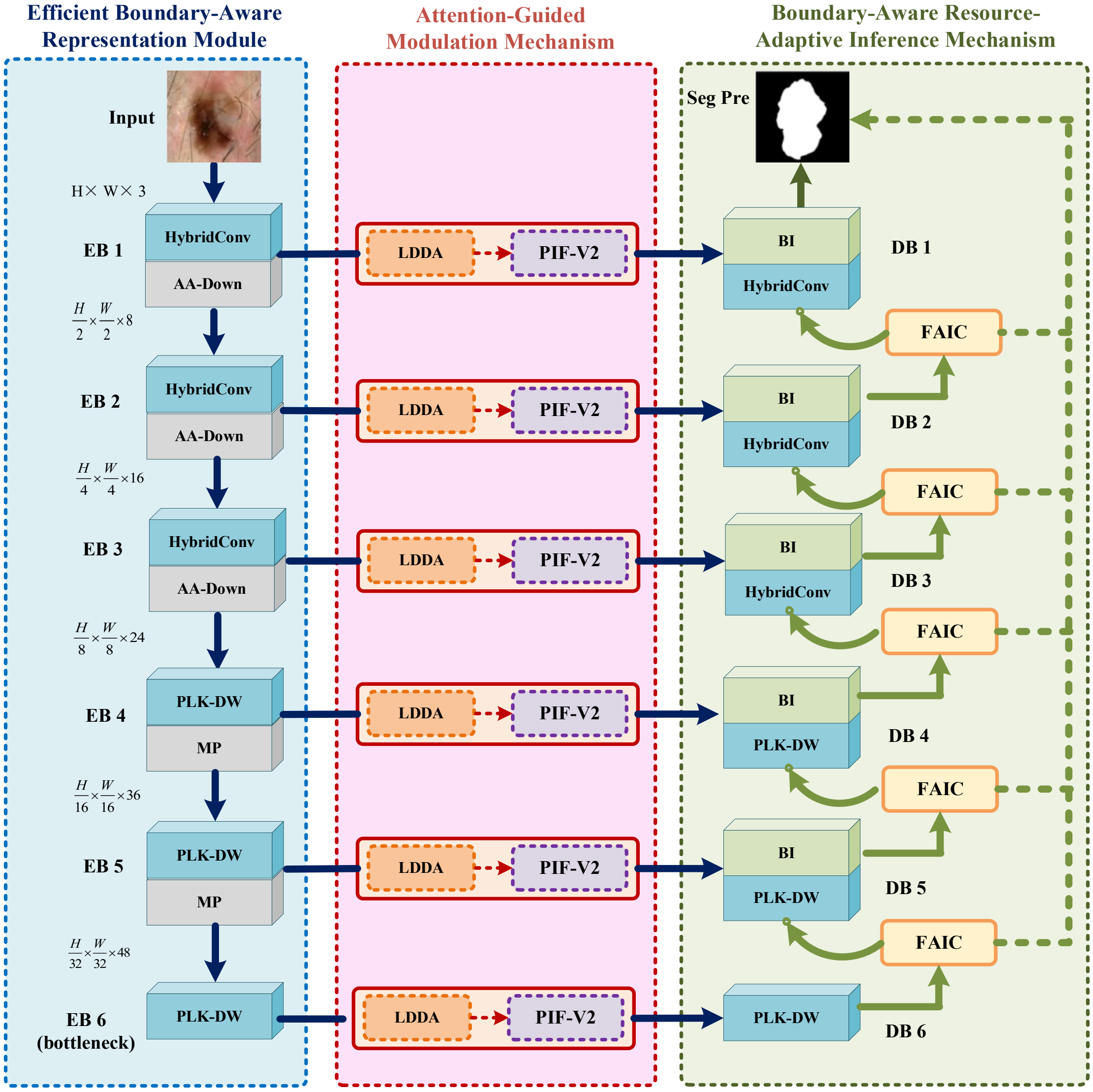

**Fig. 2**. Overall architecture of the proposed EA-LiteUNet. The network follows a U-shaped encoder–decoder structure with stage-specific convolutional design. The encoder integrates the efficient boundary-aware representation module (EBARM), skip connections are enhanced by the attention-guided modulation mechanism, and the decoder incorporates a boundary-aware resource-adaptive inference module for dynamic refinement.

Unlike conventional U-Net designs with homogeneous convolutional blocks, EA-LiteUNet adopts stage-dependent operations to accommodate the distinct characteristics of shallow and deep features. Shallow stages emphasize edge-related and local structural information, where edge-adaptive downsampling is applied to mitigate aliasing-induced boundary degradation[29]. In deeper stages, standard max-pooling is retained as feature representations become increasingly semantic and less sensitive to fine-grained perturbations. Skip connections are augmented by attention-guided modulation to selectively transmit boundary-relevant features while suppressing redundant responses, thereby reducing the semantic gap between the encoder and decoder. On the decoding side, the boundary-aware resource-adaptive inference mechanism dynamically adjusts the refinement depth according to the prediction stability and the structural complexity, creating an effective balance between the segmentation accuracy and computational efficiency[12,30]. Through this unified and stage-specific design, EA-LiteUNet achieves robust boundary preservation within a lightweight architectural framework.

### 3.2 Efficient Boundary-Aware Representation Module (EBARM)

Accurate boundary modeling requires feature representations that preserve fine-grained structural information while remaining robust to resolution reduction and noise. Conventional convolutional encoders often implicitly treat boundary-related signals, resulting in progressive degradation during feature extraction and downsampling. To address this limitation under strict computational constraints, we design an Efficient Boundary-Aware Representation Module (EBARM) that explicitly regulates boundary-related signal responses across encoder stages.

EBARM adopts a stagewise design embedded within the encoder to accommodate the distinct characteristics of shallow and deep features. In shallow layers, boundary information is strongly associated with local intensity variations and high-frequency components, making it particularly vulnerable to both aliasing and distortion. Accordingly, EBARM prioritizes boundary preservation and structural continuity in the early stages. As the feature depth increases, the focus gradually shifts toward semantic abstraction and contextual aggregation, where boundary sensitivity becomes less dominant but structural consistency remains important. To balance the preservation of local spatial details with the need for global contextual modeling, the encoder is reorganized into two functional stages: a shallow boundary-sensitive stage (Layers 1–3) and a deep semantic-context stage (Layers 4–6), as illustrated in Fig. 3.

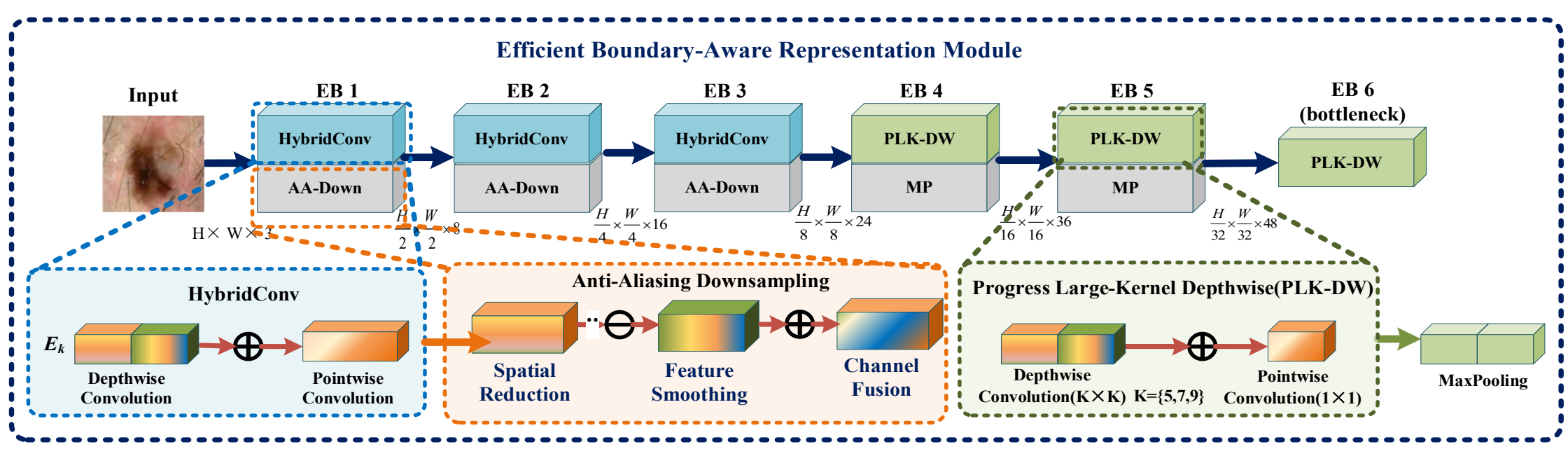

**Fig. 3.** Architecture of the efficient boundary-aware representation module (EBARM) embedded in the encoder. The encoder is divided into a shallow boundary-sensitive stage (Layers 1 – 3) and a deep semantic-context stage (Layers 4 – 6). Shallow layers employ HybridConv blocks with anti-aliasing downsampling (AA-Down) to preserve fine-grained boundary structures during early resolution reduction. Deeper layers adopt progressive large-kernel depthwise (PLK-DW) convolutions with increasing kernel sizes to enlarge the receptive field and capture global contextual information, whereas standard max-pooling is retained for semantic abstraction.

#### 3.2.1 Shallow Stage (Layers 1-3): Attention-Modulated Local Feature Extraction.

In the initial high-resolution stages, preserving fine-grained boundaries is critical. We propose the HybridConv block, integrating efficient channel attention with depthwise separable convolutions:

$$HybridConvx(x) = ECA(PWConv_{1\times1}(DWConv_{3\times3}(x))) \quad 1$$

where $DWConv_{3\times3}$ extracts spatial feature independence for each channel, $PWConv_{1\times1}$ performs channel fusion, and the efficient channel attention (ECA) module adaptively reweights channel responses to enhance boundary-relevant features while suppressing noise.

Standard strided downsampling violates the Nyquist–Shannon sampling theorem,

introducing aliasing and loss of shift-invariance that is particularly detrimental to boundary-related high-frequency information. We therefore replace standard downsampling in Layers 1–3 with Anti-Aliasing Downsampling (AA-Down), which inserts a low-pass BlurPool filter before subsampling:

$$AA - Down(x) = PWConv_{1\times1}(BlurPool(DWConv_{3\times3,s=2}(x))) \quad 2$$

Here, the stride-2 depthwise convolution performs initial spatial reduction, BlurPool suppresses high-frequency artifacts (e.g., checkerboard patterns) introduced by the strided operation, and the pointwise convolution fuses the resulting channels. This design preserves shift-invariance and provides clearer boundary guidance for subsequent stages

**3.2.2 Deep Stage (Layers 4-6): Progressive Large-Kernel Context Modeling.**

As feature resolution decreases, the modeling objective shifts from local boundary refinement to semantic abstraction and long-range context aggregation, which conventional small-kernel convolutions struggle to capture. We introduce the Progressive Large-Kernel Depthwise (PLK-DW) module, whose kernel size increases with stage depth $i \in \{4, 5, 6\}$:

$$k_i = \begin{cases} 5, i = 4 \\ 7, i = 5 \\ 9, i = 6 \end{cases} \quad 3$$

The module is defined as follows:

$$PLK - DW(x) = PWConv_{1\times1}\left(DWConv_{k_i\times k_i}(x)\right) \quad 4$$

Here, $DWConv_{k\times k}$ denotes a depthwise convolution with kernel size $k_i$, and $PWConv_{1\times1}$ denotes a pointwise convolution. This progressive expansion allows the network to adaptively enlarge its effective receptive field as the spatial resolution decreases, ensuring that the deepest layer (Stage 6) captures **global contextual information** covering the entire region of interest, while Stage 4 transitions smoothly from local to semi-global features.

### 3.3 Attention-Guided Modulation Mechanism (AGMM)

Conventional skip connections transmit encoder features to the decoder without selective filtering, allowing boundary-irrelevant or redundant responses to interfere with reconstruction. To address this limitation, we propose an attention-guided modulation mechanism. To address this limitation, we propose an attention-guided modulation mechanism that integrates a lightweight dual-domain attention (LDDA) module with a prediction information fusion strategy (PIF-V2). Rather than treating skip connections as passive feature pathways, the proposed mechanism actively regulates information flow from the encoder to the decoder via complementary attention and prediction-guided refinement strategies. Fig. 4 illustrates an overview of the proposed modulation framework.

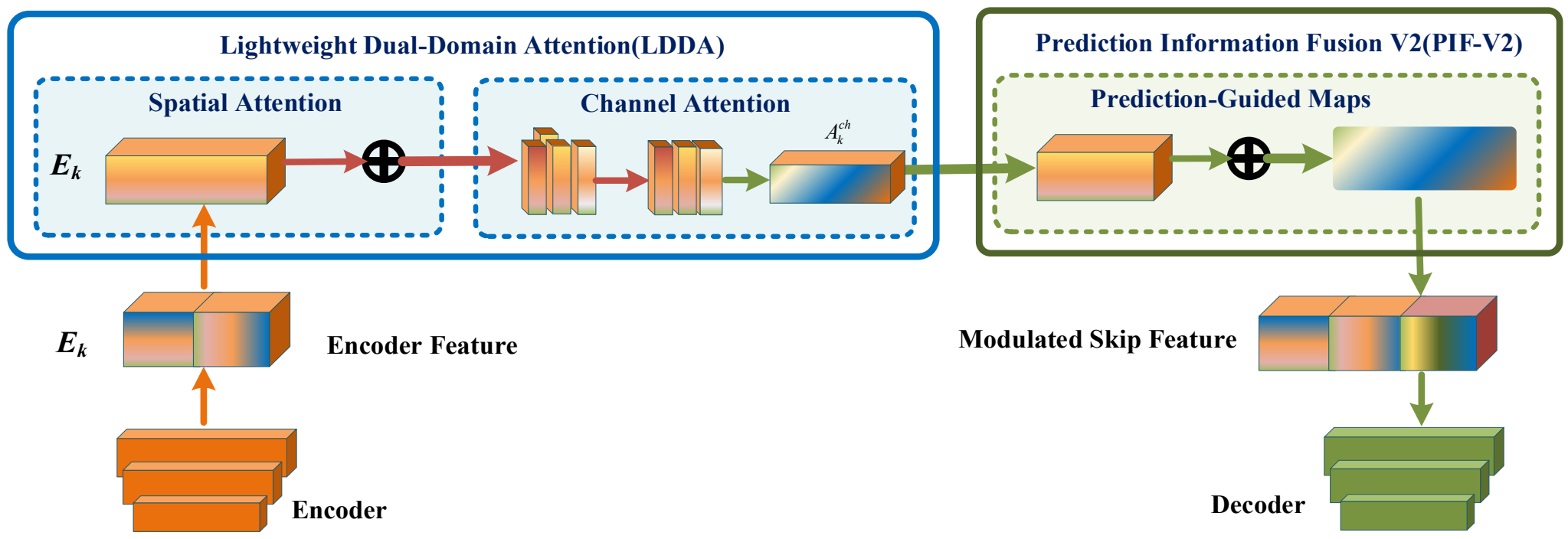

**Fig. 4.** Overview of the proposed Attention-Guided Modulation Mechanism for skip connections. The framework consists of two complementary components: a Lightweight Dual-Domain Attention (LDDA) module and a Prediction Information Fusion module (PIF-V2). LDDA sequentially applies channel-wise and spatial attention to refine encoder features by enhancing boundary-relevant responses and suppressing background noise. PIF-V2 further incorporates prediction-derived cues to adaptively regulate feature transfer, enabling selective and context-aware information propagation from the encoder to the decoder.

### 3.3.1 Lightweight Dual-Domain Attention (LDDA)

As illustrated in Fig. 4, LDDA sequentially performs channel-wise and spatial attention to regulate feature responses relevant to boundary delineation. Let $F_i \in \mathbb{R}^{B\times C\times H\times W}$ denote the encoder feature at skip-connection stage $i$.

**(1) Channel-first attention: ECA (Efficient Channel Attention Formula)**

The channel-wise attention is computed using Efficient Channel Attention (ECA) as follows:

$$w_i = \sigma(Conv1\mathrm{D}(GAP(F_i), k)), w_i \in \mathbb{R}^{B\times C} \quad 5$$

where: global average pooling is defined as:

$$\mathrm{GAP}(F_i)_{b,c} = \frac{1}{HW}\sum_{x=1}^{H}\sum_{y=1}^{W} F_{i,b,c,x,y} \quad 6$$

The resulting channel weights $w$ are broadcast and applied to the input feature map:

$$\tilde{F}_i = w_i \otimes F_i \quad 7$$

where $\otimes$ denotes channel-wise scaling. This operation selectively enhances channels encoding boundary-relevant cues while suppressing redundant responses.

**(2) Spatial-next attention**

Following channel refinement, a lightweight spatial mask is generated from the **channel-refined feature** $\tilde{F}_i$, ensuring that spatial gating operates after, and builds upon, channel selection:

$$S_i = \sigma\left(DWConv_{3\times 3}(\tilde{F}_i)\right), S_i \in \mathbb{R}^{B\times C\times H\times W} \quad 8$$

The spatially modulated feature map is then obtained as:

$$\hat{F}_i = S_i \odot \tilde{F}_i \quad 9$$

where $\odot$denotes element-wise multiplication. This step suppresses localized noise and enhances structurally salient regions, using the channel-refined feature from Eq. (7) rather than the raw input.

**(3) Dual-domain coupling: channel-then-spatial**

Substituting Eq. (7) into Eq. (9) gives the closed-form expression of the full sequential coupling, which we use for reference in subsequent sections:

$$F_i'' = S_i \odot (w_i \otimes F_i) \qquad 10$$

Since $w_i \otimes F_i = \tilde{F}_i$ by Eq. (7), $F_i''$in Eq. (10) is algebraically identical to $\hat{F}_i$in Eq. (9); Eq. (10) is provided as a single self-contained expression of LDDA's output for use in Eq. (11). By sequentially applying channel reweighting and spatial gating, LDDA jointly regulates "what" features are important and "where" they are important, thereby producing boundary-refined encoder representations suitable for skip fusion. Importantly, the use of depthwise convolution and ECA ensures that the additional computational overhead remains minimal.

**3.3.2 Prediction Information Fusion (PIF-V2)**

While LDDA refines encoder features in the feature domain, it does not explicitly account for the evolving segmentation predictions. To incorporate prediction-level guidance, we introduce prediction information fusion (PIF-V2), as shown in Fig. 4.

PIF-V2 integrates intermediate decoder predictions into the skip modulation process to provide boundary-aware feedback. Let $E_i$denote encoder features at stage $i$, and $D_i$denote decoder features. The fusion output $V_{i-1}$is defined as follows:

$$V_{i-1} = D_i + E_i + w_1^i E_i \cdot \sigma\left(\frac{\hat{R}_i}{\tau_g}\right)^{\gamma} + w_2^i E_i \cdot \sigma\left(\frac{\hat{B}_i}{\tau_g}\right)^{\gamma} \qquad 11$$

where $\hat{R}_i$ and $\hat{B}_i$ denote region and boundary-related attention maps derived from intermediate predictions, $\sigma(\cdot)$ is the sigmoid activation function, and $w_1^i, w_2^i$ represent learnable fusion weights. The temperature parameter $\tau_g$ and exponent $\gamma \in [0.5,2]$ control the sharpness and sensitivity of the adaptive gating mechanism.

By incorporating prediction-derived cues, PIF-V2 enables context-aware feature modulation, reinforcing boundary-relevant responses while maintaining lightweight computation. This prediction-guided refinement complements LDDA and enhances the reliability of skip-based information transfer.

**3.3.3 Attention-Guided Skip Modulation**

By combining LDDA and PIF-V2, the proposed attention-guided modulation mechanism transforms skip connections from passive feature transmitters into actively regulated information pathways. The mechanism first refines encoder features in the feature domain and then adaptively integrates prediction-aware cues to guide boundary-sensitive feature fusion. The additional computational overhead introduced by the two modules remains marginal, preserving the lightweight design of EA-LiteUNet-CD.

**3.4 Boundary-Aware Resource-Adaptive Inference Mechanism**

In practical medical image segmentation, boundary complexity varies significantly across samples and spatial regions. Some images contain smooth and well-defined contours, while others exhibit irregular, blurred, or ambiguous boundaries. However, conventional U-shaped networks adopt fixed-depth decoding pathways, and treat all inputs uniformly. This strategy may lead to redundant computations for boundary-stable samples or insufficient refinement for boundary-complex cases. To address this issue, we propose a boundary-aware resource-adaptive inference mechanism that dynamically regulates decoding depth according to boundary uncertainty and structural complexity. The overall structure of the proposed boundary-aware resource-adaptive inference mechanism is illustrated in Fig. 5. The proposed mechanism dynamically regulates decoder depth according to boundary-related uncertainty and structural complexity. An entropy-based confidence evaluation enables early termination for boundary-stable samples. For ambiguous cases, a boundary complexity score, which is computed from the prediction uncertainty, boundary confidence, and attention sparsity, determines whether deeper decoding stages are activated. This adaptive strategy prevents over-refinement of simple cases while allocating additional computations to boundary-complex regions.

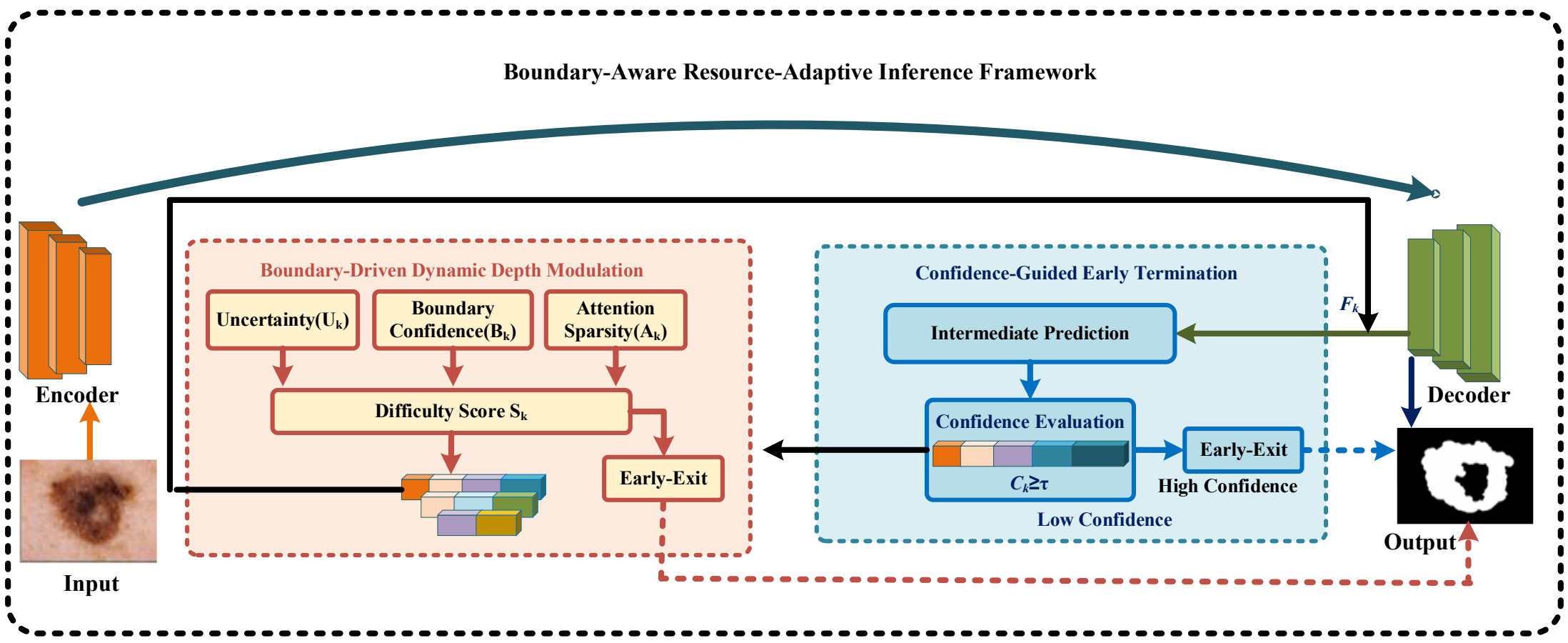

**Fig. 5.** Boundary-Aware Resource-Adaptive Inference Framework. The framework consists of two complementary components: a boundary-driven dynamic depth modulation module and a confidence-guided early termination module. The framework can dynamically regulate decoding depth according to boundary uncertainty and structural complexity.

### 3.4.1 Confidence-Guided Early-Exit Prediction Heads (EEPH)

We attach lightweight prediction heads to intermediate decoder stages to enable provisional segmentation outputs during inference. Let $\hat{Y}_k$ denote the intermediate prediction at decoder stage $k$. The prediction confidence is estimated using pixel-wise entropy:

$$C_k = 1 - \frac{1}{HW}\sum_{i,j} H(\hat{Y}_k(i,j)) \tag{12}$$

Where the binary entropy function is defined as:

$$H(p) = -p\log p - (1-p)\log(1-p) \tag{13}$$

Lower entropy indicates more stable and reliable predictions, particularly along boundary regions; $C_k$ is therefore bounded in $[0,1]$, with values near 1 indicating high confidence.

Let $\tau_c \in (0,1)$ denote the early-exit confidence threshold, jointly optimized with the network via the cost-deviation term in Eq. (27) (Section 4.3). Early termination at stage $k$ is triggered when:

$$C_k \geq \tau_c \tag{14}$$

When this condition holds, the accumulated entropy reduction indicates that further refinement is unlikely to yield a meaningful improvement in boundary localization, and decoding is terminated at stage $k$; otherwise, the network proceeds to evaluate boundary complexity, as described in Section 3.4.2, to determine whether deeper refinement is warranted.

### 3.4.2 Boundary-driven Dynamic Depth Controller (BD-DDC)

If the early-exit condition (Eq. (14)) is not satisfied at stage $k$, boundary ambiguity is assumed to persist, and we further evaluate a boundary complexity score to determine whether deeper decoding should be activated. This score aggregates three complementary signals, each capturing a distinct aspect of local prediction difficulty:

**Prediction uncertainty** $U_k$ is directly obtained from the entropy term already computed in Eq. (12)–(13), avoiding redundant computation

$$U_k = 1 - C_k \tag{15}$$

**Boundary confidence** $B_k$ measures the reliability of the prediction specifically within the boundary neighborhood, rather than averaged over the entire spatial extent as in $C_k$. Let $\mathcal{N}_k$ denote the set of pixels within a fixed morphological dilation of the predicted contour of $\hat{Y}_k$ (obtained via gradient-magnitude thresholding); boundary confidence is defined as:

$$B_k = \frac{1}{|\mathcal{N}_k|} \sum_{(i,j)\in\mathcal{N}_k} \hat{Y}_k(i,j) \tag{16}$$

so that $1 - B_k$ in the aggregation below penalizes low-confidence predictions concentrated near the contour, which are more consequential for boundary accuracy than low confidence elsewhere.

**Attention sparsity** $A_k$ quantifies how concentrated the boundary-related attention is at the corresponding skip-connection stage, using the boundary attention map $\hat{B}_i$ from PIF-V2 (Eq. (11)). We measure sparsity via its normalized entropy.

$$A_k = 1 - \frac{H(\hat{B}_i)}{\log(HW)} \tag{17}$$

where $H(\cdot)$ denotes Shannon entropy over the spatial distribution of $\hat{B}_i$, normalized by its maximum possible value $\log(HW)$; a higher $A_k$ indicates attention concentrated on a structurally distinctive (e.g., sharply defined) boundary region, whereas a diffuse, low-$A_k$ attention map suggests an ambiguous or poorly localized boundary.

The boundary complexity score at stage $k$ is then defined as a convex combination of the three signals:

$$S_k = \alpha U_k + \beta(1 - B_k) + \delta A_k, \ \alpha + \beta + \delta = 1 \tag{18}$$

where $\alpha, \beta, \delta \geq 0$ are learnable scalar weights, initialized uniformly and jointly optimized with the network end-to-end; a softmax parameterization is used to enforce the simplex constraint $\alpha + \beta + \delta = 1$, ensuring $S_k$ remains bounded in $[0, 1]$ regardless of the relative scale of $U_k$, $B_k$, and $A_k$. Let $\theta$ denote the depth-activation threshold, likewise jointly optimized via Eq. (27). The inference decision is then:

$$\begin{cases} S_k < \theta & \rightarrow \text{terminate refinement,} \\ S_k \geq \theta & \rightarrow \text{activate deeper stage,} \end{cases} \tag{19}$$

This mechanism allows the decoder to allocate computational effort in proportion to boundary complexity: samples with concentrated, high-confidence boundary predictions terminate early, while ambiguous or structurally diffuse cases receive additional refinement through the deeper PLK-DW-based decoding path (Section 3.2.2).

### 3.4.3 Hardware-Aware Computational Profiling (HAP)

EEPH (Section 3.4.1) and BD-DDC (Section 3.4.2) select decoding depth for each sample from prediction confidence and boundary complexity alone, without regard to the resource limits of the deployment device. Hardware-Aware Computational Profiling (HAP) closes this gap: it imposes a hardware-conditioned upper bound on decoder depth so that the sample-adaptive depth chosen by EEPH/BD-DDC is always executed within a device-feasible latency and memory envelope.

$D = (M, T, P)$Let the decoder comprise L = 4 stages indexed by l = 1, …, L (l = 1 denotes the shallowest). For each stage we measure an on-device resource pair (t_l, a_l) directly on the target hardware rather than estimating it from FLOPs, since kernel-launch overhead and memory-access cost make latency non-proportional to FLOPs across stages of different spatial resolution. Given a decoder executed through depth d, the cumulative latency and the peak activation memory are:

$$T(d) = \sum_{l=1}^{d} t_l, \quad M(d) = \max_{1 \le l \le d} a_l \qquad 20$$

$D = (M, T, P)$Both T(d) and M(d) are non-decreasing in d — T(d) accumulates the per-stage latency (including the gate-evaluation cost of Eqs. (12)–(19)), while M(d) is a running maximum since the decoder need not retain every stage's activation simultaneously. Given a device profile Π = (T̄, M̄) specifying a latency deadline and a memory budget, the hardware-admissible depth d_max^HW(Π) is the largest d for which T(d) ≤ T̄ and M(d) ≤ M̄, obtained by a single forward scan over the pre-profiled cost table. HAP projects the sample-adaptive depth d_sample(x) chosen by EEPH/BD-DDC onto this bound, so that the depth actually executed is:

$$d * (x; \Pi) = \min\{d_{\text{sample}}(x), \quad d_{\max}^{\text{HW}}(\Pi)\} \qquad 21$$

Because d*(x; Π) ≤ d_sample(x) for every input, HAP can only reduce, never increase, the depth selected by the sample-adaptive controller: the device budget in Π is never violated, while the learned early-exit and depth-activation behavior is left unmodified whenever the device is not the binding constraint. Table 8 (Section 4.5.3) reports the resulting accuracy-latency trade-off measured on the RTX 4060 GPU and CPU deployment targets used in this study.

# 4. Experiment

## 4.1 Datasets

The proposed model is evaluated on three dermoscopic image segmentation datasets. First, the ISIC2018 dataset[31], which contains 2594 dermoscopic images. Second, to further examine the robustness to variations in acquisition devices and lesion appearance, the PH2 dataset[32], which contains 200 dermoscopic images. Third, to enable fairer comparisons with earlier studies (e.g., U-Net++ [25], Attention U-Net[18]), we introduce the ISIC2017 dataset[33], which contains 2000 dermoscopic images. Each dataset was divided into training, validation, and test sets at a ratio of 8:1:1.

## 4.2 Evaluation Metrics

To comprehensively evaluate model segmentation accuracy, boundary quality, and computational efficiency, the Dice similarity Coefficient, intersection over union (IoU), Sensitivity (Sen), and Specificity (Spe) are used to jointly characterize lesion overlap accuracy, detection capability, and background discrimination. These metrics are defined based on the counts of true positives (TP), false positives (FP), true negatives (TN), and false negatives (FN) as follows:

$$Dice = \frac{2 \mid P \cap G \mid}{\mid P \mid + \mid G \mid} = \frac{2TP}{2TP + FP + FN} \qquad 22$$

$$mIoU = \frac{1}{c}\sum_{c=1}^{c}\frac{TP}{TP + FP + FN} \quad 23$$

$$Sensitivity = \frac{TP}{TP + FN} \quad 24$$

$$Specificity = \frac{TN}{TN + FP} \quad 25$$

The Hausdorff Distance measures the maximum boundary discrepancy between the prediction and ground truth. Formally, let $P$ and $G$ denote the set of boundary points of the prediction and the ground truth, respectively. Then, the HD95 is defined as follows:

$$HD95(P,G) = max\{h_{95}(P,G), h_{95}(G,P)\}, h_{95}(A,B) = percentile_{95}\left(\min_{b\in B}\|a-b\|\right)_{a\in A} \quad 26$$

where $\|.\|$ represents the Euclidean distance. The function $h_{95}(A,B)$ computes the 95th percentile of the directed distances from set $A$ to set $B$. We utilize the 95th percentile instead of the standard Hausdorff Distance to eliminate the impact of small outliers and noise, ensuring a more robust evaluation of the lesion contours. A lower HD95 value indicates a closer boundary match and higher segmentation precision.

In addition to accuracy metrics, model efficiency is assessed using the number of parameters (Parameters) and computational complexity measured in GFLOPs. Parameters denote the total number of trainable weights, while GFLOPs represent the number of floating-point operations required for a single forward pass with an input resolution of 256 × 256 and a batch size of 1. These metrics explicitly reflect the trade-off between segmentation performance and computational efficiency.

**4.3 Implementation details**

EA-Lite-UNet was implemented in PyTorch and trained on an NVIDIA GeForce RTX 4060 GPU. All the input images were resized to 256×256 and normalized before training. Data augmentation, including random rotation and flipping, photometric perturbations (brightness and contrast jitter), and cutout-based regularization, was applied to improve the robustness and generalizability to simulate diverse clinical scenarios.

Training was conducted with a batch size of 8 using the AdamW optimizer with an initial learning rate of $1\times10^{-3}$. A CosineAnnealingLR schedule was employed to progressively decay the learning rate. The segmentation objective combined the binary cross-entropy and Dice loss. A boundary-aware loss was adopted for boundary supervision. The resource-adaptive inference controller was optimized using a cost deviation loss to enforce the target computational budget. The total loss is expressed as follows:

$$\mathcal{L} = \mathcal{L}_{seg} + \lambda_b\mathcal{L}_{bnd} + \lambda_c\mathcal{L}_{cost} \quad 27$$

where $\lambda_b$ and $\lambda_c$ control the contributions of boundary supervision and FLOPs regulation, respectively. The resource-adaptive inference controller was trained with an additional cost-deviation term to satisfy prescribed FLOPs budgets. Specifically, target budgets were set to $\{0.5,0.75,1.0\}\times C_{\text{U-Net}}$, and the controller adjusted the early-exit and depth-activation thresholds to match the target cost. The results under multiple budgets are reported to characterize the accuracy–efficiency trade-off. Unless otherwise stated, the hyperparameters were tuned with the validation set of ISIC2018 and reused for the other datasets.

**4.4 Comparisons with *state-of -the-art* methods**

We conducted experimental evaluations of our proposed method against several state-of-the-art (SOTA) methods across three datasets. To validate the experimental approach, CNN-based networks and Mamba networks were selected.

**4.4.1 Results on the ISIC2018 Dataset**

The experimental results on the ISIC2018 dataset in Table 1 shows that our proposed EA-LiteUNet outperforms existing segmentation models across multiple key metrics, achieving the best results in terms of Dice score (92.08%), HD95 (12.89 pixels), and sensitivity (94.67%). In comparison, LiteMamba-Bound shows similar performance in terms of Dice score (90.83%) and HD95 (17.68 pixels), but slightly lower sensitivity (91.26%). While Ultralight VM-UNet achieves excellent FLOPs (0.06 G), its Dice score is notably lower (88.52%), and its sensitivity is the lowest (86.08%), indicating weaker performance in capturing edge details and small target regions compared with our model. These results demonstrate that EA-LiteUNet can better identify the target regions while maintaining high overall accuracy.

**Table 1.** Comparative experimental results on the ISIC2018 dataset. (Bold indicates the best).

| Model | Year | params(M) | FLOPs(G) | Dice (%) | SPE (%) | SEN (%) | HD95 (Px) |
|---|---|---|---|---|---|---|---|
| U-Net[28] | 2015 | 31.04 | 54.73 | 87.51±0.21 | 95.09±0.12 | 88.97±0.13 | 34.86±0.23 |
| Attention-Unet[18] | 2018 | 34.91 | 66.62 | 88.49±0.16 | 95.21±0.14 | 89.69±0.14 | 23.73±0.14 |
| UNeXt[6] | 2022 | 1.47 | 0.57 | 89.14±0.13 | 97.08±0.13 | 86.75±0.21 | 23.65±0.25 |
| U-Lite[34] | 2023 | 0.88 | 0.75 | 89.08±0.16 | 95.87±0.19 | 89.82±0.17 | 20.05±0.23 |
| EGE-Unet[35] | 2023 | 0.053 | 0.072 | 89.23±0.14 | 94.48±0.11 | 91.16±0.14 | 20.65±0.22 |
| Ultralight VM-UNet[36] | 2024 | 0.049 | **0.06** | 89.37±0.15 | 95.81±0.15 | 90.24±0.18 | 19.81±0.16 |
| LB-Unet[14] | 2024 | **0.038** | 0.098 | 88.52±0.12 | 96.87±0.12 | 86.08±0.15 | 21.41±0.17 |
| TinyUNet[37] | 2024 | 0.48 | 1.66 | 89.49±0.17 | **97.18±0.13** | 92.88±0.16 | 17.68±0.27 |
| LiteMamba-Bound[38] | 2025 | 0.96 | 1.73 | 88.43±0.15 | 94.98±0.14 | 88.06±0.19 | 18.96±0.23 |
| H-vmunet[39] | 2025 | 8.97 | 0.74 | 90.83±0.13 | 96.97±0.15 | 91.26±0.12 | 20.75±0.21 |
| MAFUNet[40] | 2025 | 9.61 | 7.43 | 89.71±0.14 | 96.95±0.12 | 88.86±0.13 | 19.16±0.13 |
| EA-LiteUNet (Ours) | 2026 | 0.29 | 1.17 | **92.08±0.09** | 97.02±0.11 | **94.67±0.12** | **12.89±0.08** |

As shown in Fig. 6, Rows 4 and 6 contain large and irregular lesion regions. Our model better preserves boundary integrity, producing segmentation results that are more closely aligned with the ground truth. In contrast, other methods, such as Att-UNet and HvmUNet, exhibit redundant edge artifacts. Rows 3 presents cases with small lesion regions, where our model accurately captures the shape characteristics of these lesions and produces smooth, natural segmentation boundaries. In contrast, some comparison methods, such as Attention-UNet and U-Lite, yield relatively rough segmentation results for these small targets. In Rows 1 and 5, which contain dark lesion regions, our model is strongly robust and accurately delineates lesion contours. In contrast, some competing methods, such as UNeXt and EGE-UNet, result in slight over-segmentation when handling these high-contrast lesions. The image in Row 2 shows sparse hair interference, but our model is effective in segmentation, demonstrating strong stability and accuracy, while other models, such as UNet and U-Lite, exhibit insufficient segmentation, and LB-Unet shows excessive segmentation.

Overall, the proposed model demonstrates strong boundary integrity. It effectively handles lesions of varying sizes and remains robust under complex background conditions. As a result, the predicted segmentation masks are closer to the ground truth annotations.

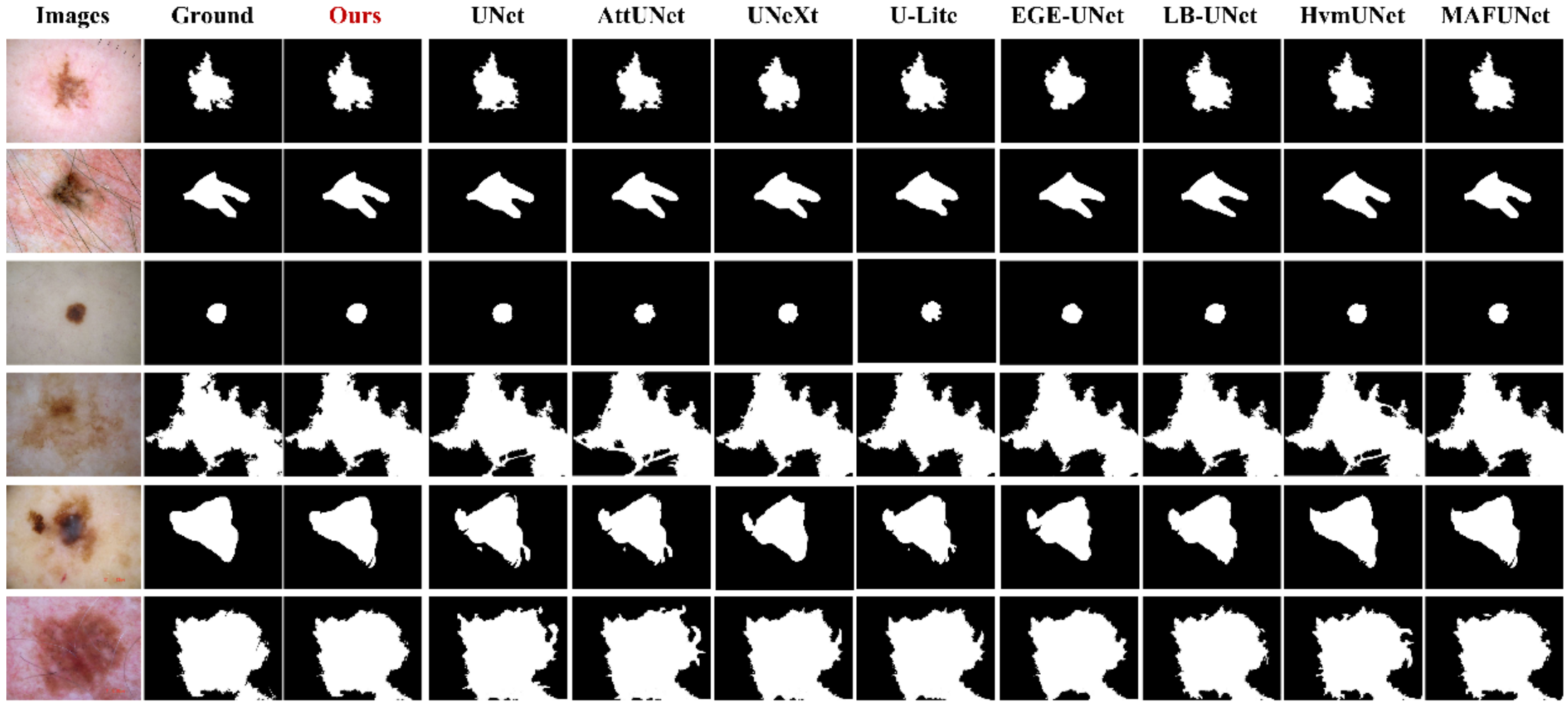

**Fig. 6.** Visual comparison with other models on the ISIC2018 dataset.

### 4.4.2 Results on the PH2 Dataset

The quantitative results on the PH2 dataset are reported in Table 2. EA-LiteUNet achieves the best overall performance across several key metrics, including the Dice score (94.17%), HD95 (12.01 pixels), and sensitivity (95.07%). Compared with the high-performing LB-UNet, our model yields modest yet consistent improvements, increasing the Dice score by 0.18%, and sensitivity by 0.26%, and reducing the HD95 by 4.93 pixels. These results indicate an enhanced ability to capture fine-grained lesion details. Compared with recent lightweight architectures, such as H-vmUNet and MAFUNet, EA-LiteUNet employs slightly more parameters but demonstrates clear advantages in terms of Dice score, HD95, and sensitivity. Compared with MAFUNet, although our model improved only the Dice score and sensitivity by 0.46% and 0.21%, respectively, the HD95 decreased from 18.16 pixels to 12.01 pixels, indicating that our model significantly improved the accuracy and stability of the segmentation boundaries and could more accurately fit the contours of real lesions.

**Table 2.** Comparative experimental results on the PH2 dataset. (Bold indicates the best).

| Model | Year | params(M) | FLOPs(G) | DSC(%) | SPE(%) | SEN(%) | HD95(Px) |
|---|---|---|---|---|---|---|---|
| U-Net[28] | 2015 | 31.04 | 54.73 | 91.34±0.13 | 94.89±0.16 | 92.97±0.13 | 32.75±0.13 |
| Attention-Unet[18] | 2018 | 34.91 | 66.62 | 92.01±0.16 | 95.01±0.12 | 93.09±0.14 | 22.03±0.15 |
| UNeXt[6] | 2022 | 1.47 | 0.57 | 91.34±0.18 | 95.98±0.21 | 93.78±0.17 | 21.85±0.12 |
| U-Lite[34] | 2023 | 0.88 | 0.75 | 91.23±0.15 | 96.87±0.19 | 92.82±0.17 | 18.05±0.17 |
| EGE-Unet[35] | 2023 | 0.053 | 0.072 | 92.18±0.14 | 95.78±0.15 | 92.16±0.11 | 19.75±0.11 |
| Ultralight VM-UNet[36] | 2024 | 0.049 | **0.06** | 92.94±0.15 | 96.01±0.12 | 93.78±0.18 | 18.81±0.16 |
| LB-Unet[14] | 2024 | **0.038** | 0.098 | 93.02±0.11 | 96.17±0.13 | 93.98±0.12 | 19.41±0.17 |
| TinyUNet[37] | 2024 | 0.48 | 1.66 | 93.99±0.21 | **97.98±0.14** | 94.81±0.12 | 17.08±0.18 |
| LiteMamba-Bound[38] | 2025 | 0.96 | 1.73 | 93.63±0.15 | 95.98±0.15 | 93.76±0.16 | 18.06±0.15 |
| H-vmunet[39] | 2025 | 8.97 | 0.74 | 93.83±0.21 | 96.27±0.21 | 93.26±0.14 | 19.05±0.14 |
| MAFUNet[40] | 2025 | 9.61 | 7.43 | 93.71±0.18 | 97.86±0.11 | 94.86±0.11 | 18.16±0.13 |
| EA-LiteUNet (Ours) | 2026 | 0.29 | 1.17 | **94.17±0.07** | 97.61±0.08 | **95.07±0.10** | **12.01±0.09** |

As shown in Fig. 7, the first column presents the original dermoscopic images, the second column shows the corresponding ground truth annotations, and the remaining columns display the segmentation results produced by the different models. Overall, our model generates segmentation results that closely match the ground truth across most cases. The lesion boundaries in Row 1 are severely blurred;

nevertheless, our model produces segmentation results that closely align with the ground truth. In the third and fourth rows, which exhibit highly irregular lesion shapes, our method effectively captures lesion shape characteristics, whereas EGE-UNet, HvmUNet, and U-Lite exhibit slight over-segmentation. For larger lesions in Rows 2 and 6, our model more accurately captures fine-grained lesion details and yields more precise boundary delineation. In contrast, the lesion in Row 5 presents a relatively regular shape, for which all the models achieve near-perfect agreement with the ground truth. Overall, these qualitative results indicate that our model achieves more stable and accurate segmentation performance than the competing methods do.

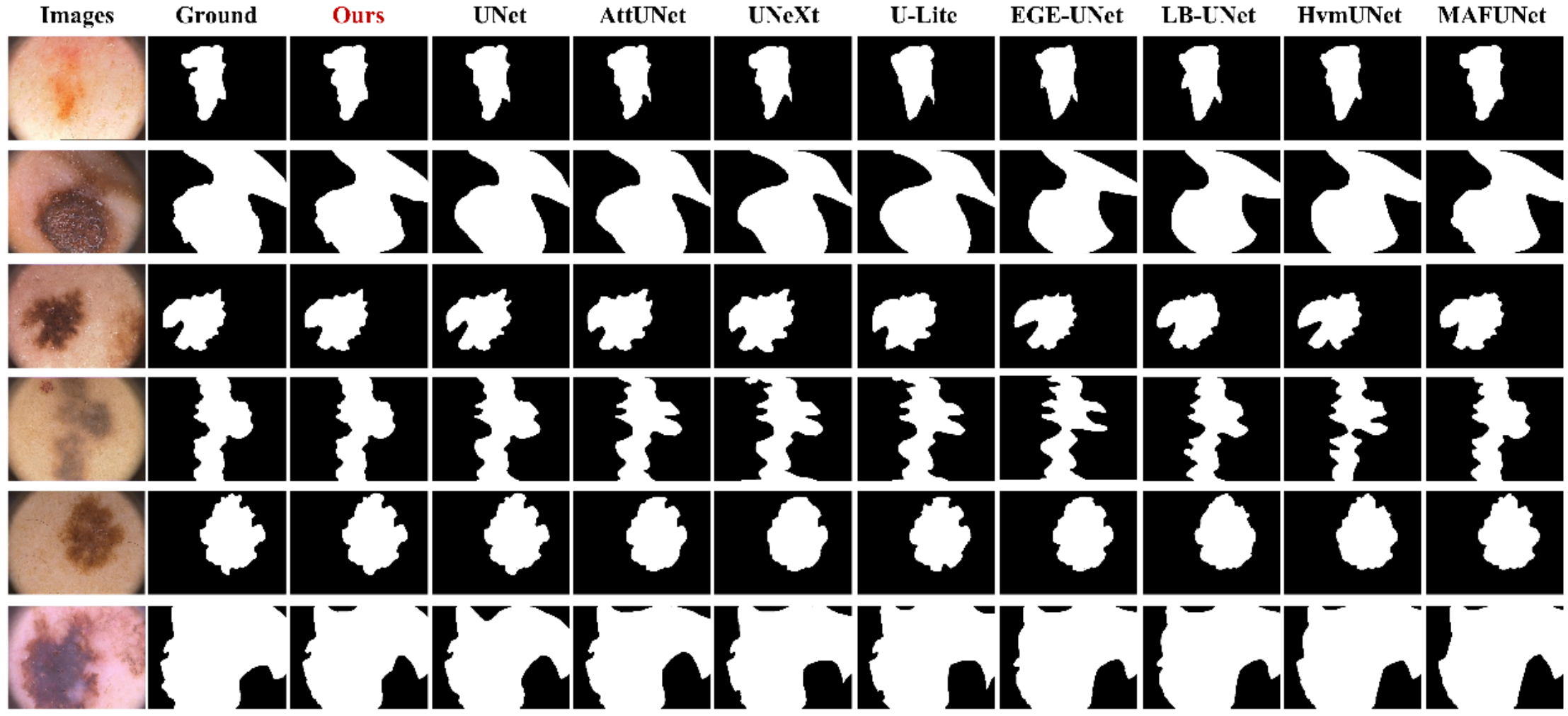

**Fig. 7.** Visual comparison with other models on the PH2 dataset.

### 4.4.3 Results on the ISIC2017 Dataset

As shown in Table 3, on the ISIC2017 dataset, EA-LiteUNet achieves competitive performance, with a Dice score of 89.12%, an HD95 of 14.23 pixels, and sensitivity of 93.07%, outperforming most existing methods. Compared with LB-UNet, which also demonstrates strong performance (Dice of 88.85%, HD95 of 18.57 pixels, and sensitivity of 92.75%), EA-LiteUNet yields modest yet consistent improvements in Dice coefficient and sensitivity of 0.27% and 0.21% while HD95 is reduced by 4.34 pixels, indicating an enhanced ability to capture segmentation boundary details.

EA-LiteUNet further improves the accuracy by 0.36% over EGE-UNet and achieves a lower HD95 than larger models such as Attention-Unet and U-Net do, suggesting more effective suppression of false-positive segmentations. Although its specificity (95.89%) is slightly lower than that of the lightweight LB-UNet (96.01%), the overall gains in Dice, HD95, and sensitivity result in more balanced segmentation performance.

**Table 3**. Comparative experimental results on the ISIC2017 dataset. (Bold indicates the best).

| Model | Year | params(M) | FLOPs(G) | DSC(%) | SPE(%) | SEN(%) | HD95(Px) |
|---|---|---|---|---|---|---|---|
| U-Net[28] | 2015 | 31.04 | 54.73 | 86.84±0.17 | 94.35±0.18 | 90.17±0.16 | 36.75±0.13 |
| Attention-Unet[18] | 2018 | 34.91 | 66.62 | 87.01±0.15 | 95.01±0.16 | 90.59±0.17 | 25.23±0.14 |
| UNeXt[6] | 2022 | 1.47 | 0.57 | 87.89±0.14 | 95.98±0.17 | 91.12±0.12 | 24.65±0.15 |
| U-Lite[34] | 2023 | 0.88 | 0.75 | 88.12±0.17 | 96.17±0.12 | 89.76±0.16 | 20.78±0.23 |
| EGE-Unet[35] | 2023 | 0.053 | 0.072 | 88.15±0.16 | 95.38±0.12 | 91.96±0.11 | 21.91±0.12 |
| Ultralight VM-UNet[36] | 2024 | 0.049 | **0.06** | 88.76±0.12 | 95.86±0.16 | 91.89±0.13 | 20.16±0.16 |
| LB-Unet[14] | 2024 | **0.038** | 0.098 | 88.92±0.13 | 95.17±0.14 | 91.91±0.14 | 22.41±0.15 |
| TinyUNet[37] | 2024 | 0.48 | 1.66 | 88.85±0.14 | **96.01±0.13** | 92.75±0.15 | 18.57±0.19 |
| LiteMamba-Bound[38] | 2025 | 0.96 | 1.73 | 87.88±0.19 | 95.98±0.15 | 91.26±0.16 | 19.06±0.18 |
| H-vmunet[39] | 2025 | 8.97 | 0.74 | 89.23±0.18 | 96.27±0.13 | 92.01±0.14 | 21.05±0.14 |
| MAFUNet[40] | 2025 | 9.61 | 7.43 | 88.21±0.15 | 95.26±0.17 | 89.87±0.13 | 20.76±0.13 |
| EA-LiteUNet (Ours) | 2026 | 0.29 | 1.17 | **89.52±0.12** | 95.89±0.13 | **89.07±0.12** | **14.23±0.11** |

As shown in Fig. 8, the image in Row 2 contains hair interference and irregular lesion shapes. Our model accurately delineates lesion contours and produces smooth, natural segmentation boundaries. In contrast, U-Net and EGE-UNet exhibit irregular edges and slight over-segmentation in these cases. Rows 4 present lesions with more complex morphologies and uneven color distribution. Under such challenging conditions, our method maintains reliable segmentation accuracy, whereas U-Net, Att-UNet, and HvmUNet tend to suffer from under-segmentation. Row 5 shows large, irregular lesions, for which our model effectively preserves overall lesion morphology and boundary characteristics. By comparison, U-Lite and LB-UNet demonstrate noticeable deviations along lesion edges. In Row 3, where the lesion shape is relatively regular, our model precisely localizes the lesion area and achieves segmentation results that closely match the ground truth. However, U-Lite and UNeXt exhibit varying degrees of over-segmentation. Overall, these qualitative results indicate that our model achieves robust and accurate segmentation across lesions of different sizes, morphologies, and background complexities.

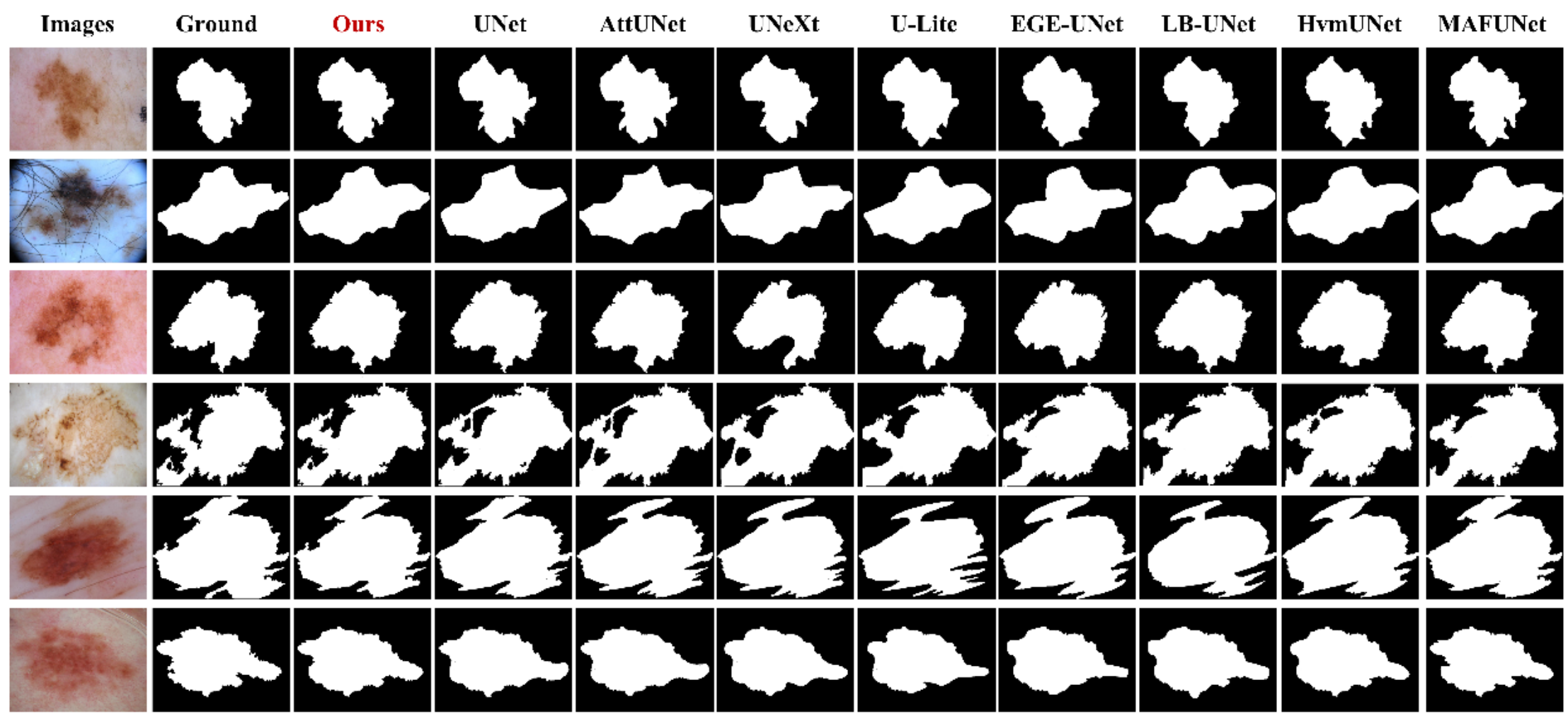

**Fig. 8.** Visual comparison with other models on the ISIC2017 dataset.

### 4.5 Ablation Study

### 4.5.1 Module-Level Ablation

#### *4.5.1.(a) Cumulative ablation study on ISIC-2018*

To systematically evaluate the effectiveness of each proposed component in EA-LiteUNet, ablation experiments were conducted on the ISIC2018 dataset under identical training settings, including backbone configuration, optimizer, learning schedule, and data augmentation strategy. The Dice score and HD95 were adopted to measure region-based overlap and boundary accuracy, respectively. The quantitative results are summarized in Table 4.

**Table 4**. Ablation study results on the ISIC2018 dataset.

| Model | HybridConv | AA-Down | PLK-DW | LDDA | PIF-V2 | Adaptive Inference | Dice(%) | HD95(Px) |
|---|---|---|---|---|---|---|---|---|
| Model1 | | | | | | | 85.61±0.11 | 19.89±0.12 |
| Model2 | ✓ | | | | | | 85.98±0.12 | 18.94±0.10 |
| Model3 | ✓ | ✓ | | | | | 86.43±0.13 | 16.35±0.11 |
| Model4 | ✓ | ✓ | ✓ | | | | 88.87±0.10 | 15.78±0.09 |
| Model5 | ✓ | ✓ | ✓ | ✓ | | | 90.86±0.12 | 14.62±0.13 |
| Model6 | ✓ | ✓ | ✓ | ✓ | ✓ | | 91.59±0.11 | 13.23±0.12 |
| Model7(Ours) | ✓ | ✓ | ✓ | ✓ | ✓ | ✓ | 92.08±0.09 | 12.89±0.08 |

The final segmentation performance of the progressively augmented configurations is reported in Table 4. Starting from the baseline model, the introduction of HybridConv yields simultaneous improvements in the Dice score and HD95, indicating enhanced local structural representation. Replacing conventional downsampling with anti-aliasing downsampling (AA-Down) leads to a more substantial reduction in HD95, confirming that suppressing aliasing stabilizes high-frequency boundary signals during early feature extraction. Incorporating the PLK-DW module further increases the Dice score, reflecting the improved semantic coherence enabled by enlarged receptive fields. Together, these components constitute the Efficient boundary-aware representation module (EBARM).

Embedding LDDA into skip connections consistently lowers HD95, suggesting that selective feature modulation suppresses boundary-irrelevant activations. The addition of PIF-V2 further refines boundary localization, particularly in ambiguous or low-contrast regions. Finally, the integration of adaptive inference results in additional HD95 reduction with minimal variation in Dice, indicating that dynamic refinement primarily benefits structurally complex contours without compromising region consistency. Overall, the cumulative improvements validate the complementary and non-redundant contributions of the proposed modules.

***4.5.1.(b) Training Dynamics Analysis***

To further investigate the impact of each component during optimization, we analyze the training trajectories of the Dice score and HD95 across epochs.

**(1) Region-Based Convergence (Dice)**

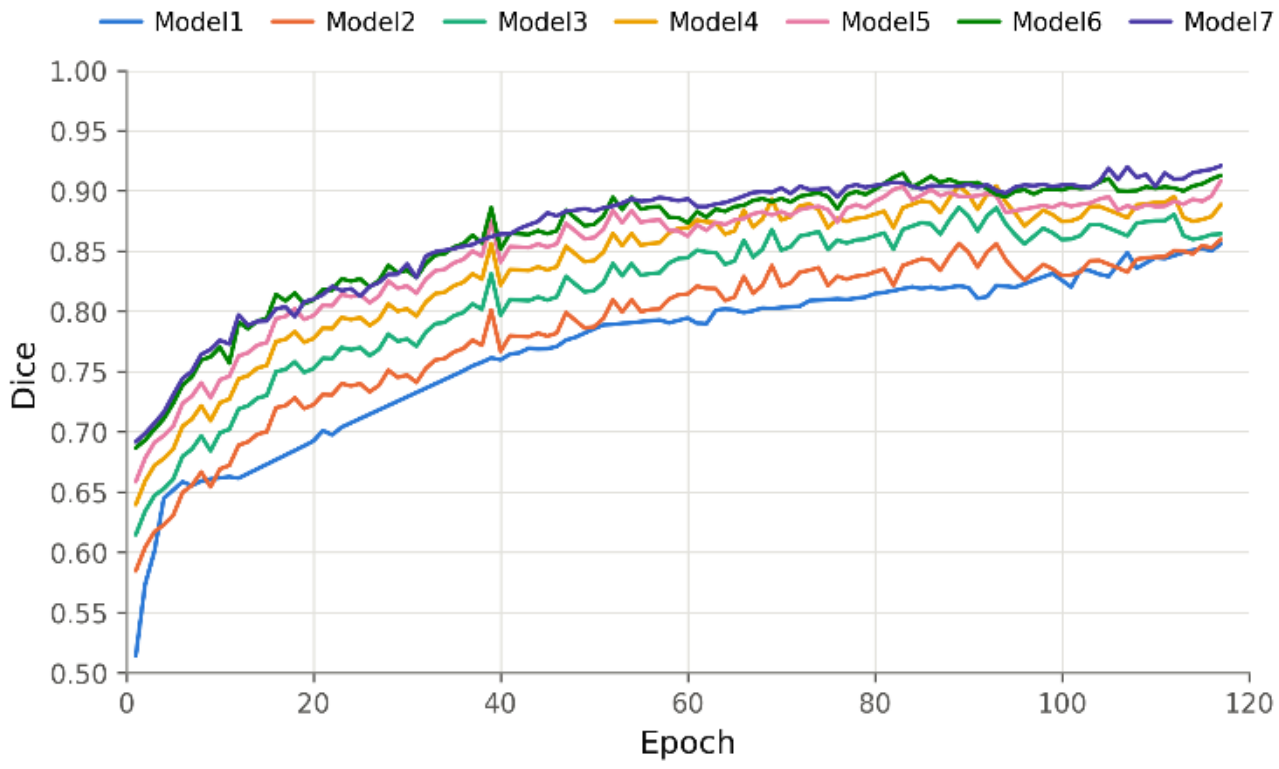

**Fig. 9.** Dice vs. Epoch for the ablation study. Evolution of the Dice scores across training epochs for different ablation configurations. Dashed lines indicate knee epochs, while dotted lines denote convergence phases. The full configuration achieves higher Dice values with smoother trajectories and earlier stabilization

The Dice evolution curves are illustrated in Fig. 9. As the modules are progressively incorporated, the Dice trajectories shift upward and exhibit a smoother convergence behavior. The model-1 (baseline) configuration shows larger oscillations and slower stabilization, reflecting an unstable region prediction. In contrast, model-7 (full configuration) achieves higher Dice values with reduced variance and earlier convergence, as evidenced by the earlier knee epoch and stable plateau phase. These observations indicate that representation-level stabilization and attention-guided modulation improve regional consistency during training.

**(2) Boundary Refinement Dynamics (HD95)**

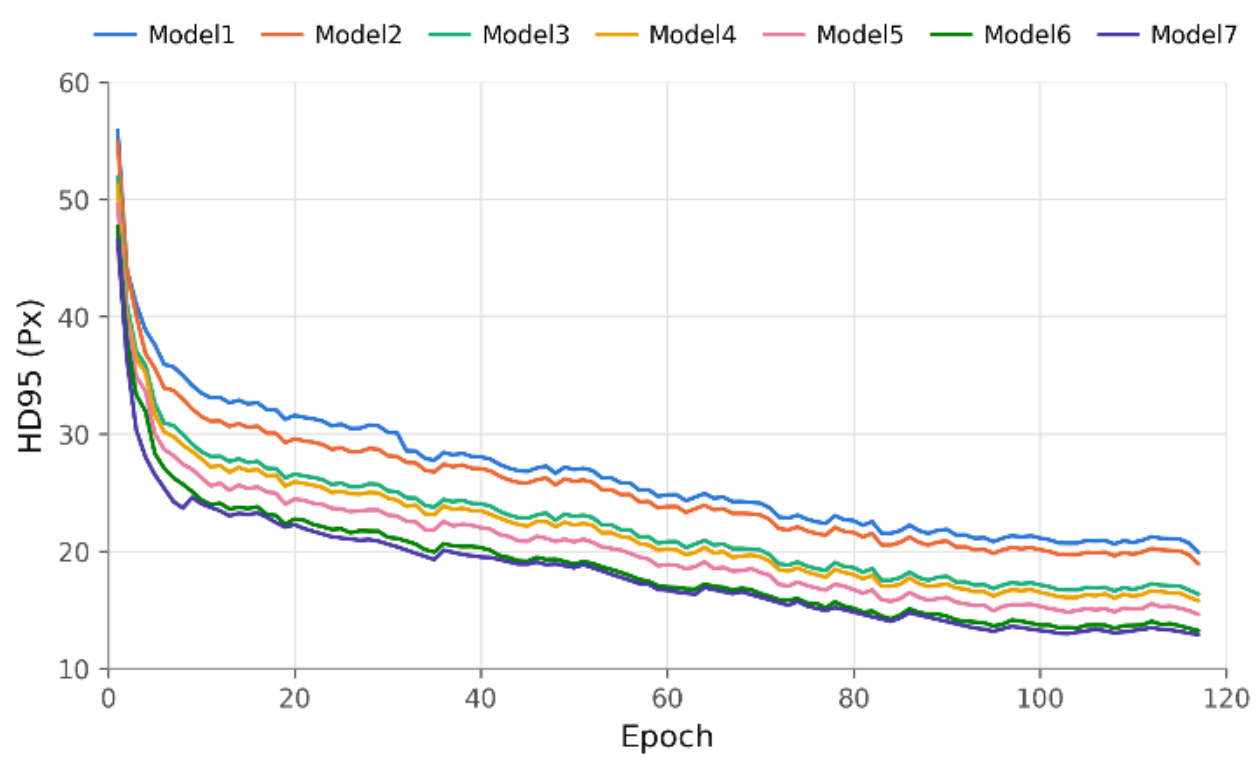

**Fig. 10.** HD95 vs. Epoch for the ablation study. Evolution of the HD95 across training epochs for different ablation configurations. The progressive downward shift of the curves indicates a cumulative boundary refinement. The full configuration results in faster early-stage reduction and lower final convergence values

Fig. 10 presents the corresponding HD95 curves. A consistent downward shift is observed as additional modules are integrated. Model-1 (baseline) has higher boundary error and larger fluctuations, indicating limited robustness in contour modeling. Model-4 (Intermediate configurations) shows progressively reduced HD95 and improved stability. Notably, Model-7 (full configuration) exhibits faster early-stage reduction in the HD95 and lower final convergence, suggesting improved coarse boundary localization and finer contour precision. The reduced oscillation amplitude further indicates an improved structural stability during optimization.

**4.5.2 Orthogonal Ablation of LDDA and PIF-V2**

In the cumulative ablation (Table 4), LDDA and PIF-V2 are introduced sequentially (Model4→Model5→Model6), so the measured contribution of PIF-V2 is conditioned on LDDA already being present. This leaves open the possibility that the improvement attributed to PIF-V2 partly reflects additional optimization the model would have undergone regardless—i.e., that PIF-V2's apparent gain is confounded with LDDA still being under-trained at the point PIF-V2 is introduced, rather than reflecting PIF-V2's own contribution. To disentangle the two effects, we conduct a fully orthogonal ablation with four configurations—neither, LDDA-only, PIF-V2-only, and both—starting from the EBARM-only backbone (Model4) on ISIC2018.

**Table 5.** Orthogonal ablation of LDDA and PIF-V2 on ISIC2018

| Model | LDDA | PIF-V2 | Dice(%) | HD95(Px) |
|---|---|---|---|---|
| Base (Model4) | | | 88.87±0.10 | 15.78±0.09 |
| Base+LDDA | ✓ | | 91.06±0.15 | 14.62±0.11 |
| Base+PIF-V2 | | ✓ | 89.68±0.13 | 13.58±0.10 |
| Base+LDDA+PIF-2(Model6) | ✓ | ✓ | 91.59±0.11 | 13.23±0.12 |

As in the cumulative ablation, LDDA and PIF-V2 are each individually beneficial relative to the EBARM-only backbone (Model4) on both datasets, and their joint configuration (Model6) achieves both the best Dice/HD95 and the lowest run-to-run variance among the four configurations on both datasets, consistent with the smoother convergence behavior observed for the full model in Section 4.5.1 (Figs.

9–10). Independent of the precise magnitude of this interaction, the orthogonal ablation confirms the result motivating its design: PIF-V2 yields a clear improvement over Model4 even in the complete absence of LDDA (+0.81 Dice, −2.20px HD95), ruling out the possibility that its contribution in the cumulative ablation (Table 4) merely reflects an under-trained LDDA at the point it is introduced.

### 4.5.3 Sub-Components Level Ablation

#### *4.5.3.(a) EBARM Internal Components*

Table 6 decomposes EBARM into its three constituent sub-components—AA-Down, HybridConv, and PLK-DW —introduced in Section 3.2.1–3.2.2, and ablates each in turn from the complete module, holding AGMM and BA-RAIM fixed. This sub-component leave-one-out protocol is applied consistently to AGMM (Table 7) and BA-RAIM (Table 8) in the following two subsections, isolating each module's internal design choices from its macro-level contribution already established in Section 4.5.1.

**Table 6.** EBARM Internal Components Ablation

| Configuration | Params (M) | FLOPs (G) | Dice (%) | mIoU (%) | HD95(px) | SPE(%) | SEN(%) |
|---|---|---|---|---|---|---|---|
| Full EBARM (AA-Down + HybridConv + PLK-DW) | 0.29 | 1.17 | 92.08±0.09 | 84.74±0.14 | **12.89±0.08** | **97.02±0.11** | 92.67±0.12 |
| w/o AA-Down (S1-S3: Standard $\text{Conv}_{\text{stride-2}}$) | 0.28 | 1.15 | 91.82±0.11 | 84.51±0.12 | 15.37±0.14 | 96.85±0.12 | 92.34±0.13 |
| w/o HybridConv (S1-S3: Standard $\text{Conv}_{3\text{x}3}$) | 0.35 | 1.38 | 91.55±0.13 | 84.23±0.15 | 15.34±0.16 | 96.62±0.13 | 92.08±0.14 |
| w/o PLK-DW (S4-S6: Standard $\text{Conv}_{3\text{x}3}$) | 0.31 | 1.22 | 91.68±0.12 | 84.35±0.14 | 14.79±0.15 | 96.71±0.12 | 92.15±0.13 |

***Note:*** *Evaluated on the ISIC2018 test set (Section 4.4.1), the first row reproduces the complete EBARM in Table 4 (the complete Model7/Ours configuration). PLK-DW replaces the standard fixed-kernel convolution in Stages 4–6 with a progressively expanding depthwise kernel (5×5→7×7→9×9), reflecting the increasing need for long-range context modeling as spatial resolution decreases. The "w/o PLK-DW" configuration substitutes a fixed 3×3 depthwise convolution across all three stages, isolating the contribution of progressive kernel expansion itself.*

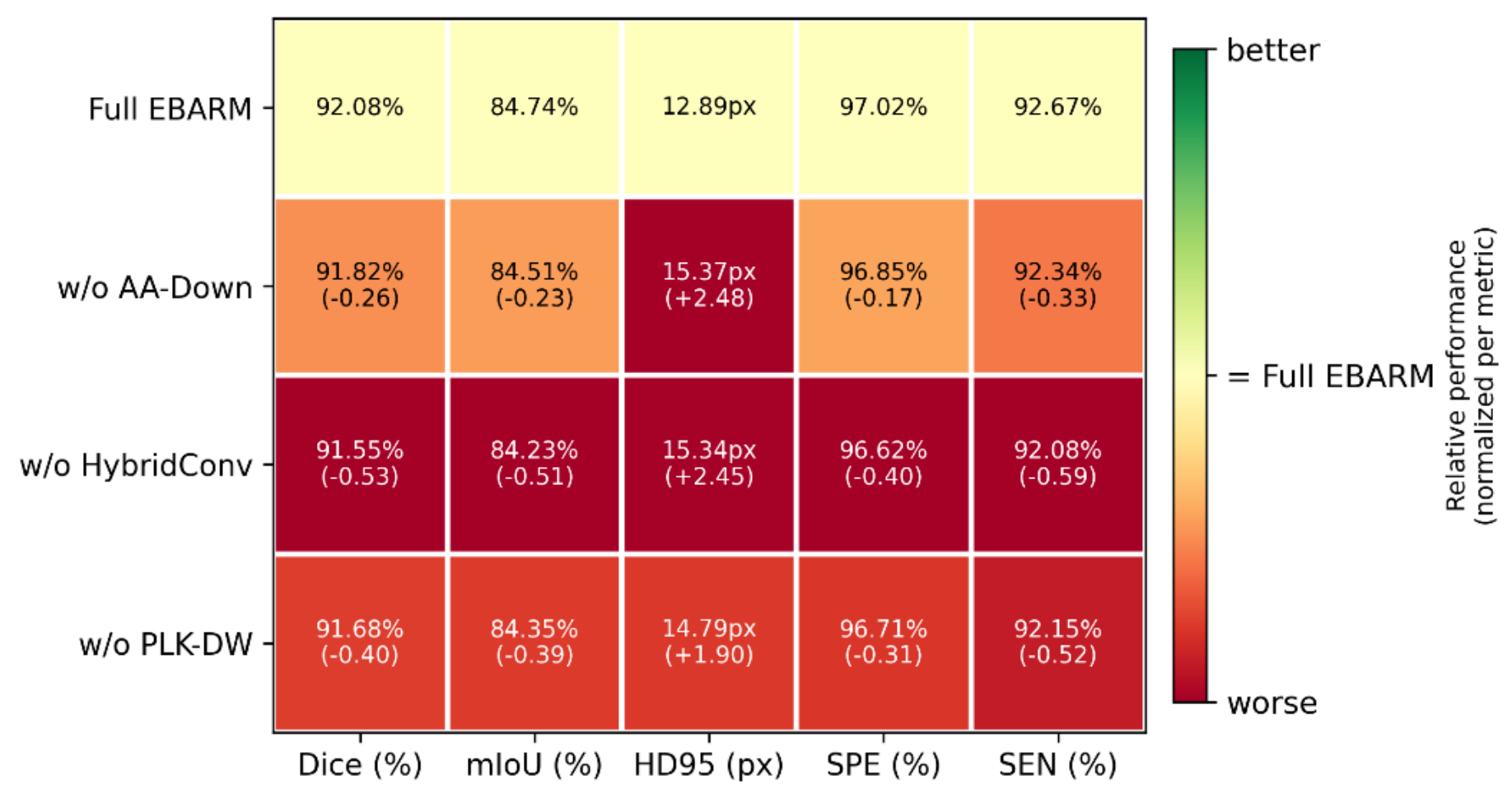

**Fig. 11.** EBARM Internal Components Ablation: Relative Performance Heatmap. Rows are the Table 6 configurations, columns are the five evaluation metrics; color encodes each cell's value relative to Full EBARM (green = better, red = worse), normalized per metric, with the raw value and Δ from Full EBARM annotated in each cell.

HybridConv proves the most broadly critical: its removal produces the largest degradation on four of the five metrics (ΔDice = −0.53, ΔmIoU = −0.51, ΔSPE = −0.40, ΔSEN = −0.59), consistent with its role as the primary local feature extractor in the shallow stages (Section 3.2.1) — replacing it degrades the representation that every downstream component, including AGMM and BA-RAIM, depends on. AA-Down, by contrast, shows a boundary-specific profile: it costs the least in Dice of the three (−0.26) yet produces the single largest HD95 degradation (+2.48px), confirming that its anti-aliasing downsampling specifically protects boundary localization during spatial reduction (Section 3.2.1) rather than contributing to overall region overlap. PLK-DW has the smallest impact of the three on every metric (ΔHD95 = +1.90px, the smallest of the three), consistent with its role in the deep stages (S4–S6): operating at lower spatial resolution, its progressively expanding receptive field mainly aids semantic context aggregation rather than fine-grained boundary precision, so its removal is comparatively less costly to the boundary-sensitive metrics this thesis prioritizes. Together, these results identify HybridConv as the component EBARM cannot do without, while AA-Down's disproportionate HD95 role — a large boundary-accuracy cost paired with only a modest Dice cost — indicates that boundary preservation within EBARM is concentrated in a single specialized mechanism rather than distributed evenly across all three sub-components.

***4.5.3.(b) AGMM Internal Components***

Table 7 decomposes AGMM into its two constituent sub-mechanisms — PIF-V2 and LDDA, introduced in Section 3.3 — and ablates each in turn from the complete module, holding EBARM and BA-RAIM fixed. Rows 4–6 further decompose LDDA into its channel and spatial attention steps, and PIF-V2 into its learnable gating parameters (τ, γ), and are therefore nested within, rather than orthogonal to, Rows 2–3. This sub-component leave-one-out protocol follows the same structure used for EBARM (Table 6).

**Table 7.** AGMM Internal Components Ablation

| Configuration | Params(M) | FLOPs(G) | Dice (%) | mIoU (%) | HD95(px) | SPE(%) | SEN(%) |
|---|---|---|---|---|---|---|---|
| Full AGMM (PIF-V2+LDDA) | 0.29 | 1.17 | 92.08±0.09 | 84.74±0.14 | **12.89±0.08** | **97.02±0.11** | 92.67±0.12 |
| w/o PIF-V2 | 0.29 | 1.17 | 91.62±0.12 | 84.31±0.14 | 15.87±0.15 | 96.68±0.12 | 92.12±0.13 |
| w/o LDDA | 0.29 | 1.17 | 91.45±0.13 | 84.12±0.15 | 16.45±0.16 | 96.52±0.13 | 91.89±0.14 |
| w/o Channel Attention | 0.29 | 1.17 | 91.78±0.11 | 84.48±0.13 | 14.56±0.14 | 96.82±0.12 | 92.34±0.13 |
| w/o Spatial Attention | 0.29 | 1.17 | 91.85±0.11 | 84.55±0.13 | 13.92±0.13 | 96.89±0.12 | 92.41±0.13 |
| w/o *τ*, *γ* | 0.29 | 1.17 | 91.72±0.12 | 84.38±0.14 | 14.78±0.14 | 96.75±0.12 | 92.18±0.13 |

***Note:*** *Evaluated on the ISIC2018 test set (Section 4.4.1). EBARM and BA-RAIM held fixed; Row 1 reproduces Table 4 (the complete Model7/Ours configuration). Rows 2–3 ablate AGMM's two components (PIF-V2, LDDA); Rows 4–6 further decompose LDDA (channel/spatial) and PIF-V2 (learnable τ, γ), and are thus nested within Rows 2–3, not orthogonal to them.*

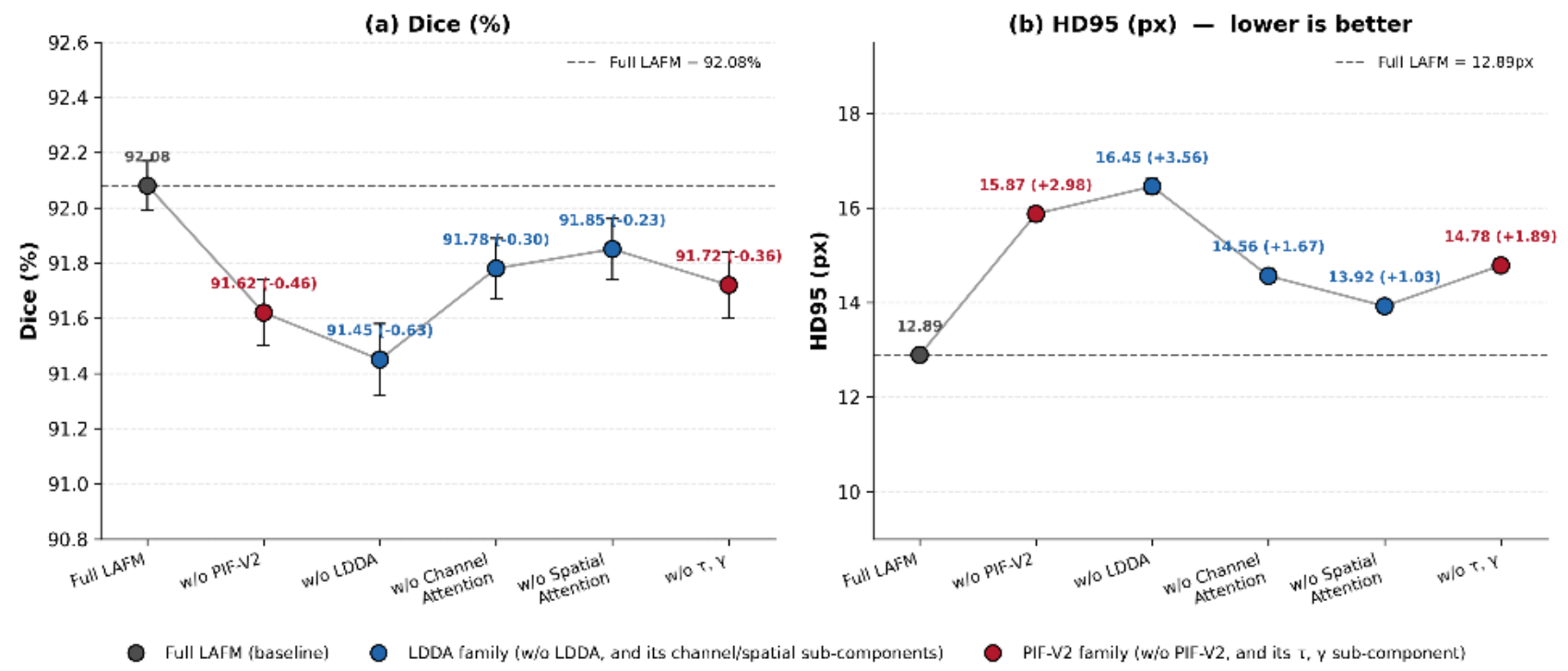

**Fig. 12.** AGMM Internal Components Ablation. (a) Dice (%) and (b) HD95 (px) for each Table 7 configuration; the dashed line marks the Full AGMM value, and color distinguishes the LDDA family (w/o LDDA and its channel/spatial sub-components) from the PIF-V2 family (w/o PIF-V2 and its τ, γ sub-component)**.**

Removing either component degrades both metrics, confirming that PIF-V2 and LDDA are complementary rather than redundant. LDDA is the more critical component overall (ΔDice = −0.63, ΔHD95 = +3.56), while PIF-V2 shows a boundary-specific profile: it costs little in Dice (−0.46) but is disproportionately important for HD95 (+2.98). Within LDDA, channel attention drives most of the gain — removing it costs more on both Dice (−0.30 vs. −0.23) and HD95 (+1.67 vs. +1.03) than removing spatial attention. This matches the design rationale in Section 3.3.1: channel reweighting sharpens the feature map first, making the subsequent spatial mask easier to learn. The learnable τ and γ in PIF-V2 also matter: fixing them to a static sigmoid costs 0.36 pp in Dice and 1.89 px in HD95, confirming that adaptive gating sharpness is not a cosmetic addition. Finally, the sub-component losses do not sum linearly to the full-module loss (channel + spatial: ΔHD95 = +2.70, vs. +3.56 for LDDA as a whole), indicating a synergistic rather than additive relationship between the two attention steps. This boundary-specific profile mirrors the pattern observed for AA-Down within EBARM (Section 4.5.3(a)): in both modules, a single sub-component disproportionately protects boundary accuracy while contributing comparatively little to overall Dice, suggesting that boundary preservation in this architecture is achieved through a small set of specialised, non-redundant mechanisms distributed across EBARM and AGMM rather than through any single dominant component.

#### *4.5.3.(c) BA-RAIM Internal Components*

Table 8 decomposes BA-RAIM into its three constituent sub-mechanisms — EEPH (Section 3.4.1), BD-DDC (Section 3.4.2), and HAP (Section 3.4.3) — and ablates each in turn from the complete configuration, following the same leave-one-out protocol used for EBARM (Table 6) and AGMM (Table 7). EBARM and AGMM are held fixed throughout, isolating the individual contribution of each BA-RAIM sub-mechanism to segmentation accuracy and inference efficiency.

**Table 8.** BA-RAIM Internal Components Ablation

| Configuration | Params (M) | FLOPs (G) | Dice (%) | mIoU (%) | HD95(px) | SPE(%) | SEN(%) | Laten-GPU(ms) | Laten-CPU(ms) |
|---|---|---|---|---|---|---|---|---|---|
| Full BA-RAIM | 0.29 | 0.62 | 92.08±0.09 | 84.74±0.14 | 12.89±0.08 | 97.02±0.11 | 92.67±0.12 | 13 | 48 |
| w/o EEPH | 0.29 | 1.02 | **92.15±0.08** | **84.79±0.13** | **12.56±0.07** | **97.06±0.10** | **92.78±0.09** | 22 | 142 |
| w/o BD-DDC | 0.29 | **0.55** | 91.68±0.10 | 84.28±0.14 | 14.87±0.12 | 96.72±0.11 | 92.15±0.12 | **11** | **38** |
| w/o HAP | 0.29 | 0.68 | 91.85±0.10 | 84.52±0.14 | 13.67±0.08 | 96.88±0.11 | 92.38±0.12 | 15 | 52 |

***Note:*** *Evaluated on the ISIC2018 test set (Section 4.4.1); row 1 reproduces the complete BA-RAIM configuration in Table 4 (the complete Model7/Ours configuration). EEPH, BD-DDC, and HAP correspond to the sub-mechanisms introduced in Sections 3.4.1, 3.4.2, and 3.4.3, respectively; each row removes exactly one sub-mechanism while holding EBARM and AGMM fixed. Δ values discussed in the text are relative to the complete BA-RAIM configuration (row 1). Unlike Table 4, where unmarked FLOPs denote static peak computation, every FLOPs value reported here is the average dynamic FLOPs measured under each ablated inference configuration on the test set.*

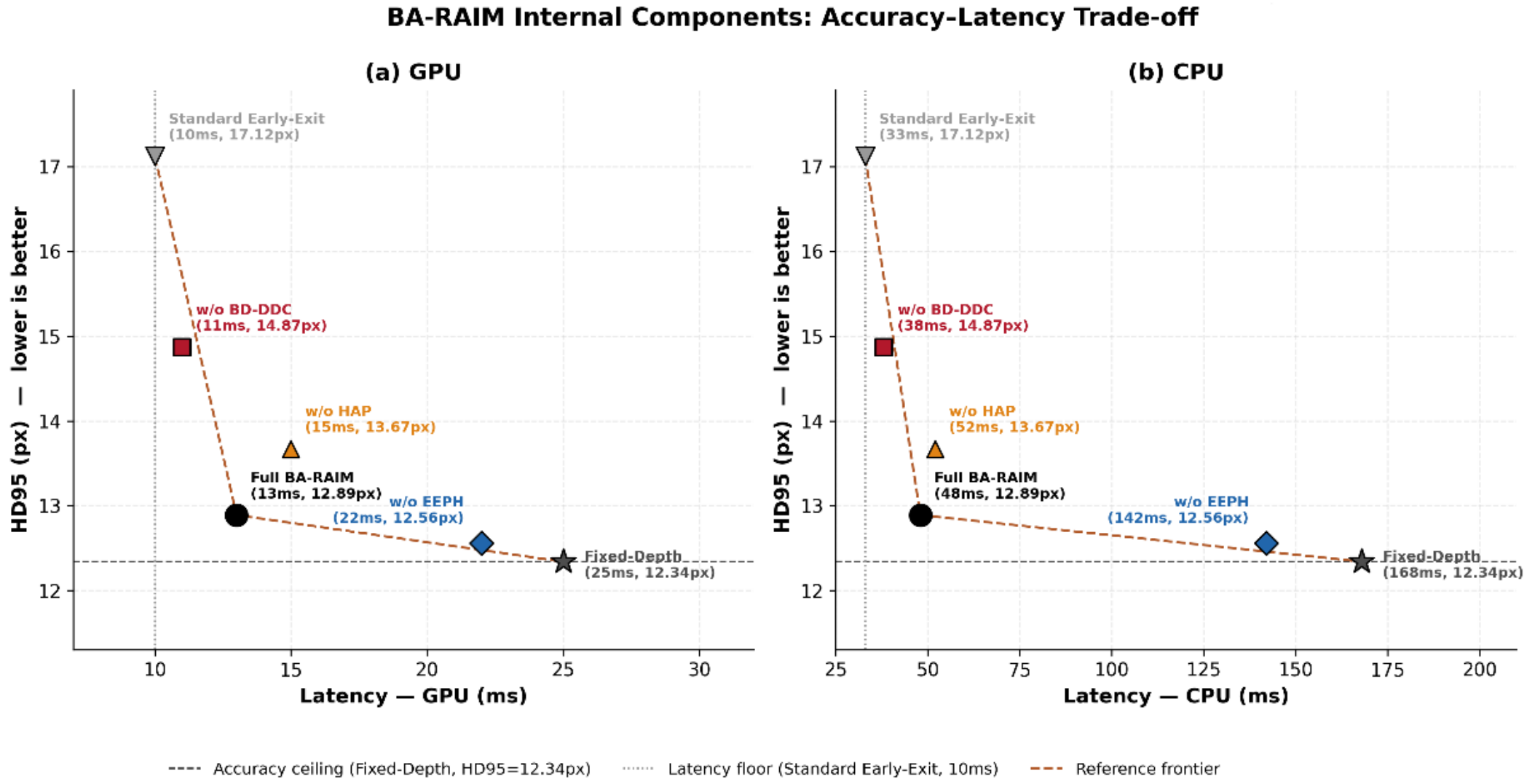

**Fig. 13.** BA-RAIM Internal Components Ablation: Accuracy-Latency Trade-off. HD95 (px) vs. latency for the Table 8 configurations plus the Fixed-Depth and Standard Early-Exit reference strategies from Table 8, measured on (a) GPU and (b) CPU. The dashed line traces the reference Pareto frontier defined by the two Table 8 strategies and Full BA-RAIM; the horizontal and vertical dotted lines mark the Fixed-Depth accuracy ceiling and the Standard Early-Exit latency floor, respectively**.**

Removing BD-DDC causes the largest accuracy loss (ΔDice = −0.40, ΔHD95 = +1.98 px, +15.4%) yet the fastest inference and lowest FLOPs (0.55G) among all configurations, confirming that its boundary-complexity-driven depth activation (Eq. (18)– (19)) is the primary safeguard against premature exit on ambiguous boundaries.

Removing EEPH leaves accuracy essentially unchanged (ΔDice = +0.07, ΔHD95 = −0.33 px) but nearly doubles FLOPs (0.62G→1.02G) and latency (+69.2% GPU, +195.8% CPU), approaching but not reaching the Fixed-Depth ceiling (Table 8). This isolates EEPH as a purely efficiency-oriented gate: without it, every sample must still pass through BD-DDC's evaluation, so some early termination remains.

Removing HAP produces a smaller, intermediate shift (ΔDice = −0.23, ΔHD95 = +0.78 px; FLOPs +9.7%; latency +2 ms GPU/+4 ms CPU), consistent with its role as a hardware-conditioned depth cap that binds only occasionally on a single fixed platform, rather than a per-sample decision criterion.

Together, the results establish a consistent ranking for accuracy protection (BD-DDC > HAP > EEPH), inverted for compute savings (EEPH > HAP > BD-DDC), confirming that early exit and depth modulation are complementary, with BD-DDC indispensable for preserving BA-RAIM's boundary-accuracy target.

## 5. Discussion

The experimental results demonstrate that EA-LiteUNet achieves consistent joint improvement in boundary-based (HD95) and region-based (Dice, Sensitivity) metrics across all three datasets, indicating that enhanced contour modeling does not come at the expense of overall segmentation overlap. This joint improvement contrasts with the pattern observed among several efficient baselines in Tables 1–3: for example, LiteMamba-Bound attains a Dice score close to that of EA-LiteUNet but with markedly higher HD95, while Ultralight VM-UNet achieves the lowest FLOPs at the cost of both Dice and sensitivity. Such trade-offs suggest that boundary fidelity and regional overlap are commonly treated as competing objectives in lightweight segmentation design; the results obtained here indicate that the two can instead be jointly optimized when boundary-related signals are explicitly regulated throughout the network, rather than left as an implicit by-product of region-based supervision alone.

The ablation study (Section 4.5) indicates that the three proposed components act in a complementary and non-redundant manner along the boundary-signal pathway: EBARM mitigates aliasing and preserves high-frequency structural information during encoding; the LDDA – PIF-V2 modulation mechanism selectively reweights skip-connection features to emphasize boundary-relevant responses while suppressing redundant ones; and BA-RAIM adaptively allocates additional decoding depth to samples with unresolved boundary ambiguity. Progressive inclusion of these components not only lowers HD95 monotonically (Section 4.5.1) but also yields smoother and earlier convergence of boundary-related metrics during training (Fig. 9–10), suggesting that regulating high-frequency boundary signals reduces gradient noise and stabilizes optimization, rather than merely improving the converged endpoint.

From a signal-processing perspective, EA-LiteUNet can be viewed as a hierarchical framework for boundary-frequency regulation. Lesion boundaries correspond to high-frequency components that are particularly vulnerable to degradation during downsampling, feature fusion, and decoding. The proposed architecture addresses this vulnerability at three levels: representation-level preservation (EBARM), feature-level selective gating (LDDA–PIF-V2), and inference-level adaptive refinement (BA-RAIM). This layered regulation offers one plausible explanation for the simultaneous improvement in boundary precision and regional accuracy observed under strict computational constraints; we note that this interpretation is motivated by empirical consistency across the ablation results rather than a formal frequency-domain derivation.

Beyond accuracy, the practical significance of EA-LiteUNet lies in its resource efficiency: achieving these improvements with only 0.29M parameters and 1.17 GFLOPs (Table 1) makes it substantially lighter than most compared CNN- and Mamba-based baselines. Combined with the resource-adaptive inference mechanism's ability to trade accuracy for further latency reduction under prescribed FLOPs budgets (Section 4.3), this efficiency suggests that the proposed design is well suited to deployment on resource-constrained clinical or point-of-care devices, where both diagnostic accuracy and inference latency are practically important.

Despite these advantages, several limitations should be acknowledged. First, validation is currently restricted to dermoscopic skin-lesion datasets (ISIC2017, ISIC2018, PH2); generalization to other imaging modalities and anatomical targets (e.g., CT, MRI, or endoscopic images) has not been verified and remains an open question. Second, the resource-adaptive inference mechanism relies on entropy-based uncertainty estimation and thresholds tuned on a single validation set (ISIC2018), which may not transfer optimally to datasets with substantially different boundary statistics. Third, latency and FLOPs budgets were benchmarked on a single GPU/CPU platform (Section 4.5.3); results may vary across hardware with different memory bandwidth and parallelism characteristics. Finally, although performance is reported with variability across evaluation runs (Tables 1 – 4), we did not perform formal statistical significance testing against baselines, nor did we obtain expert clinical evaluation of the qualitative segmentation results; both would strengthen the clinical credibility of the reported improvements.

Future work will therefore pursue four directions: (1) extending evaluation to cross-modality and multi-organ datasets to assess generalization beyond dermoscopy; (2) replacing the fixed, heuristically tuned thresholds with learnable or meta-learned decision criteria for the adaptive inference mechanism; (3) benchmarking inference efficiency across a broader range of edge and mobile hardware; and (4) conducting statistical significance testing and expert clinical evaluation to further validate the reported improvements. In parallel, a more formal frequency-domain analysis of boundary signal propagation may help substantiate the signal-processing interpretation proposed above.

**6. Conclusion**

In this work, we presented EA-LiteUNet, an edge-adaptive and resource-efficient U-shaped architecture for boundary-sensitive medical image segmentation. Motivated by the vulnerability of boundary-related high-frequency signals to degradation during downsampling, feature fusion, and decoding, the proposed framework integrates boundary-aware representation learning, attention-guided skip modulation, and adaptive inference control within a unified lightweight design.

Unlike conventional U-Net variants that primarily emphasize region-based optimization, EA-LiteUNet explicitly regulates boundary-sensitive signals throughout the segmentation pipeline: the encoder mitigates aliasing-induced structural distortion, the skip connections selectively enhance contour-relevant features, and the decoder dynamically adjusts the refinement depth according to the prediction confidence. This coordinated strategy enables simultaneous improvement in boundary fidelity and region consistency while maintaining low computational complexity.

Extensive experiments on three public dermoscopic datasets (ISIC2017, ISIC2018, PH2) demonstrate that EA-LiteUNet achieves competitive or superior performance compared with recent lightweight and attention-based baselines; on ISIC2018, for instance, it attains a Dice score of 92.08% and reduces HD95 to 12.89 pixels using only 0.29M parameters and 1.17 GFLOPs. Ablation analysis further confirms the complementary, non-redundant contribution of each proposed module to this result.

Overall, EA-LiteUNet offers a practical and resource-efficient solution for boundary-sensitive dermoscopic image segmentation, and the signal-processing perspective adopted in its design may inform future work on efficient, boundary-aware architectures for medical image analysis more broadly. Building on the limitations discussed in Section 5, future efforts will focus on cross-modality generalization, learnable adaptive-inference thresholds, and clinical validation to further establish the practical utility of the proposed framework.

**Data Availability**

The datasets analyzed during the current study are publicly available benchmark datasets:

The ISIC 2017 Dataset is available at https://challenge.isic-archive.com/data/#2017

The ISIC 2018 Dataset is available at https://challenge.isic-archive.com/data/#2018

The PH2 Dataset is available at https://www.fc.up.pt/addi/ph2%20database.html

All datasets are freely accessible for non-commercial research purposes.

**Acknowledgements**

We would like to express our sincere appreciation to our research team for their valuable contributions to this work. Their insights, feedback, and support played a crucial role in shaping the content and improving the overall quality of the manuscript. We also extend our gratitude to the researchers, authors, and publishers of the studies cited in this paper for their significant contributions to the field.

**Author contributions**

All authors contributed to the conception and design of the study. J.T.W. designed and performed the experiments and conducted the data analysis. N.I.R.R. coordinated the manuscript preparation and conducted critical revision. P.P.F. and Y.H. were responsible for data collection and preprocessing. Y.Y. verified and validated the experimental results. All authors reviewed and approved the final version of the manuscript.

**Competing Interests**

The authors declare no competing interests.

**Funding**

The authors received No Funding for this work.